\documentclass[10pt,twocolumn,letterpaper]{article}
\usepackage{cvpr}
\usepackage{times}
\usepackage{epsfig}
\usepackage{graphicx}
\usepackage{amsmath}
\usepackage{amssymb}
\usepackage{comment}
\usepackage{color}
\usepackage{enumerate}
\usepackage{bm}
\usepackage[ruled,linesnumbered,vlined]{algorithm2e}
\usepackage{multirow}
\usepackage{tabularx} 
\usepackage{booktabs}
\usepackage[font=small]{caption}
\usepackage{subcaption}
\usepackage{listings}
\usepackage{slash}
\DeclareMathOperator*{\argmin}{arg\,min}
\DeclareMathOperator{\trace}{trace}
\DeclareMathOperator{\diag}{diag}

\newcommand{\ba}{\mathbf{a}}
\newcommand{\bA}{\mathbf{A}}
\newcommand{\bb}{\mathbf{b}}
\newcommand{\bbf}{\mathbf{f}}

\newcommand{\br}{\mathbf{r}}
\newcommand{\bs}{\mathbf{s}}
\newcommand{\bt}{\mathbf{t}}

\newcommand{\bx}{\mathbf{x}}
\newcommand{\by}{\mathbf{y}}

\newcommand{\bphi}{\boldsymbol{\phi}}
\newcommand{\btheta}{\boldsymbol{\theta}}
\newcommand{\bel}{\boldsymbol{\ell}}
\newcommand{\bSigma}{\boldsymbol{\Sigma}}
\newcommand{\bC}{\mathbf{C}}

\newcommand{\bH}{\mathbf{H}}
\newcommand{\bR}{\mathbf{R}}
\newcommand{\bW}{\mathbf{W}}

\newcommand{\bmo}{\mathbf{m}}
\newcommand{\bv}{\mathbf{v}}
\newcommand{\bT}{\mathbf{T}}
\newcommand{\bM}{\mathbf{M}}
\newcommand{\bU}{\mathbf{U}}
\newcommand{\bV}{\mathbf{V}}
\newcommand{\sS}{\mathcal{S}}
\newcommand{\sT}{\mathcal{T}}

\newcommand{\plucker}{Pl$\ddot{\text{u}}$cker}
\newcommand{\bbR}{\mathbb{R}}

\newcommand{\transpose}{^\mathsf{T}}

\newcommand{\figref}[1]{Figure~\ref{#1}}
\newcommand{\eqnref}[1]{(\ref{#1})}
\newcommand{\secref}[1]{Section~\ref{#1}}

\newcommand{\algoref}[1]{Algorithm~\ref{#1}}

\SetKwInput{KwData}{Inputs}
\SetKwInput{KwResult}{Output}
\SetKwComment{Comment}{$\triangleright$\ }{}
\usepackage[pagebackref=true,breaklinks=true,letterpaper=true,colorlinks,bookmarks=false]{hyperref}

\cvprfinalcopy % *** Uncomment this line for the final submission

\def\cvprPaperID{3096} % *** Enter the CVPR Paper ID here

\ifcvprfinal\fi
\begin{document}

%%%%%%%%% TITLE
\title{\plucker Net: Learn to Register 3D Line Reconstructions}

\author{Liu Liu $^{1,2}$,  Hongdong Li $^{1,2}$, Haodong Yao $^1$~and Ruyi Zha $^1$\\ 
$^{1}$ Australian National University, Canberra, Australia  \\
$^{2}$ Australian Centre for Robotic Vision \\
\tt\small{\url{https://github.com/Liumouliu/PlueckerNet}}\\
\tt\small{Liu.Liu}@anu.edu.au
}

\maketitle
%\thispagestyle{empty}

%%%%%%%%% ABSTRACT
\begin{abstract}
Aligning two partially-overlapped 3D line reconstructions in Euclidean space is challenging, as we need to simultaneously solve correspondences and relative pose between line reconstructions. This paper proposes a neural network based method and it  
% This paper proposes a neural network based method to align two partially overlapping 3D line reconstructions in Euclidean space. Our method
has three modules connected in  sequence: (i) a Multilayer Perceptron (MLP) based network takes \plucker \ representations of lines as inputs, to extract discriminative line-wise features and matchabilities (how likely each line is going to have a match), (ii)  an Optimal Transport (OT) layer takes two-view line-wise features and matchabilities as inputs to estimate a 2D joint probability matrix, with each item describes the matchness of a line pair, and (iii) line pairs with Top-K matching probabilities are fed to a $2$-line minimal solver in a RANSAC framework to estimate a six Degree-of-Freedom ($6$-DoF) rigid transformation. Experiments on both indoor and outdoor datasets show that registration (rotation and translation) precision of our method outperforms baselines significantly.

\end{abstract}
%%%%%%%%% BODY TEXT
\section{Introduction}

\begin{figure}
\begin{center}
\includegraphics[trim=7cm 11cm 7cm 12cm, clip=true, width=0.22\textwidth]{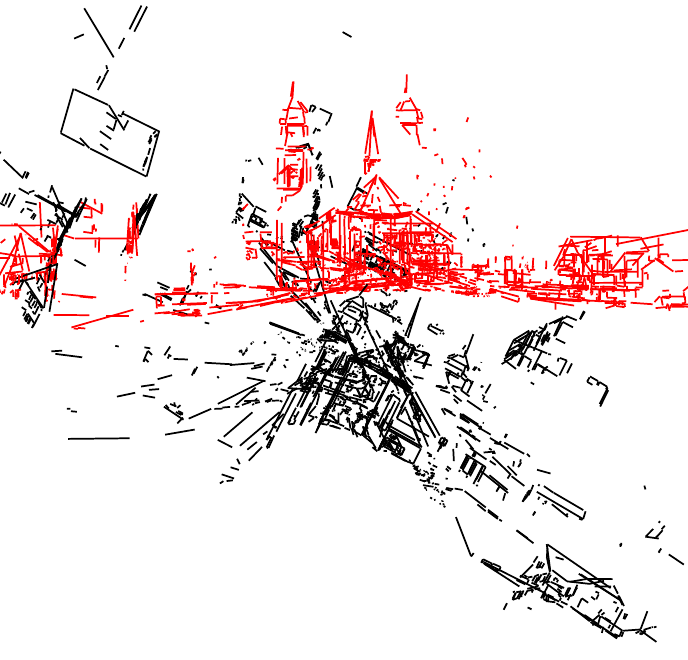}\hfill
\includegraphics[trim=7cm 11.5cm 7cm 12cm, clip=true, width=0.22\textwidth]{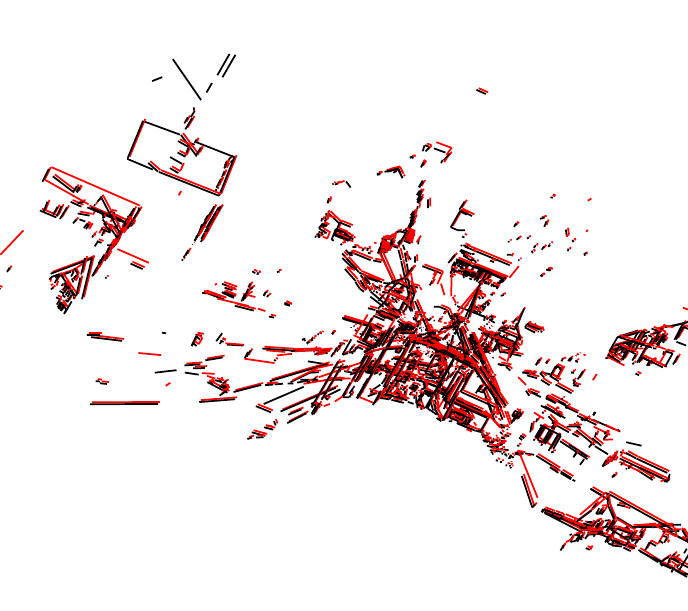}
\end{center}
% \vspace{-10pt}
\caption{ \it Our problem is to align two partially-overlapped line reconstructions or, equivalently, to estimate the relative pose between two line reconstructions. \textbf{Left}: \textcolor{red}{Red} and Black lines (depicting street-view buildings and landmarks) are in two different coordinate systems from the Semantic3D dataset \cite{hackel2017isprs}. \textbf{Right}: our method is able to successfully align the two line reconstructions in a one-shot manner.}
% \vspace{10pt}
\label{fig:problem_figure}
\end{figure}

Lines contain strong structural geometry information of environments (even for texture-less indoor scenes), and are widely used in many applications, \eg, SLAM~\cite{smith2006real,zhou2015structslam}, visual servoing~\cite{andreff2002visual}, place recognition~\cite{taubner2020lcd} and camera pose estimation~\cite{liu1990determination,lee2019elaborate}.  The underlying 3D lines can be obtained from structure from motion~\cite{taylor1995structure}, SLAM~\cite{zhang2015building} or laser scanning~\cite{ma2019generation}. Compared with 3D points, scene represented by lines is more complete and requires significantly less amount of storage~\cite{koch2016automatic,hofer2015line3d,lu2019fast}.
Given 3D line reconstructions, a fundamental problem is how to register them (\figref{fig:problem_figure}). This technique can be used in building a complete 3D map, robot localization, SLAM, \etc.

This paper studies the problem of aligning two partially-overlapped 3D line reconstructions in Euclidean space. This is not doable for traditional methods as it's very hard to find line matches by only checking 3D line coordinates, often one needs to manually set line matches \cite{bartoli20013d,bartoli2003motion} or assumes lines are mostly located on planes~\cite{koch2016automatic} and windows~\cite{cohen2016indoor}. With deep neural networks, we give a learning-based solution, dubbed as \plucker Net.

It is non-trivial to learn from lines, as we need to carefully handle line parameterization and geometry. For example, local structure defined by geometric nearest neighbor is a core-component in point-based networks (\eg, PointNet++~\cite{qi2017pointnet}).
% However, directly defining geometric nearest neighbors on line coordinates would be non-sense as there is no universally agreed error metric for comparing lines \cite{bartoli20013d}.
However, for line-based network, defining geometric nearest neighbor is non-trivial as there is no universally agreed error metric for comparing lines~\cite{bartoli20013d}.

% Based on the knowledge of traditional line geometry \cite{pottmann2009computational},
We parameterize a 3D line using a deterministic $6$-dim \plucker~\cite{pottmann2009computational} representation with a $3$-dim direction vector lying on a 3D unit hemisphere and a $3$-dim moment vector. To capture local line structure, for each line, we first extract local features in the subspace of direction and moment, and then combine them to obtain a global high-dim line feature. To make line-wise features discriminative for matching, we use a graph neural network with attention mechanism~\cite{sarlin2020superglue, simon2020stickypillars}, as it can integrate contextual cues considering high-dim feature embedding relationships.

As we are addressing a partial-to-partial matching problem, lines do not necessarily have to match. We model the likelihood that a given line has a match and is not an outlier by regressing line-wise prior matchability.  Combined with line-wise features from two line reconstructions, we are able to estimate line correspondences in a global feature matching module. This module computes a weighting (joint probability) matrix using optimal transport, where each element describes the matchability of a particular source line with a particular target line. Sorting the line matches in decreasing order by weight produces a prioritized match list, which can be used to recover the $6$-DoF relative pose between source and target line reconstructions. 

With line matches are found, we develop a $2$-line minimal pose solver (\cite{bartoli2003motion}) to solve for the relative pose in Euclidean space. We further show how to integrate the solver within a RANSAC framework using a score function to disambiguate inliers from outliers.

\begin{figure}
\begin{center}
\includegraphics[width=0.49\textwidth]{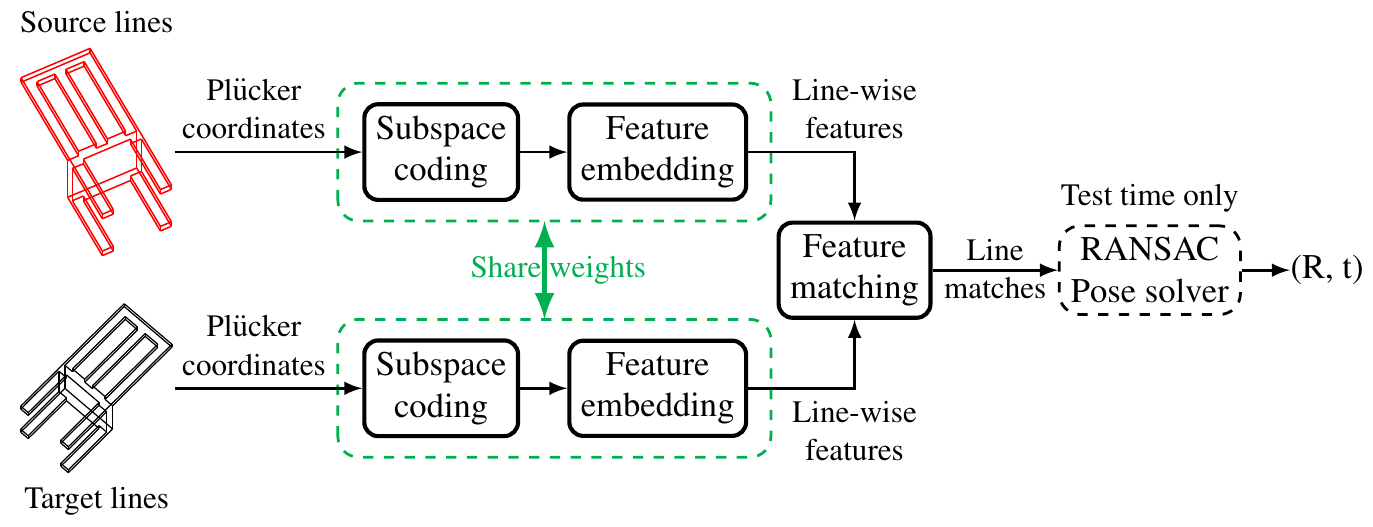}
\end{center}
% \vspace{-10pt}
\caption{\it Overall pipeline of our method.
First, lines are represented as $6$-dim \plucker \ coordinates, passed into a Siamese network to extract line-wise deep features via subspace coding and discriminative feature embedding (\secref{sec:feature_extraction}).
Then a global feature matching module estimates line matches from these features using an optimal transport (OT) technique~\cite{villani2009optimal,cuturi2013sinkhorn,courty2016optimal} (\secref{sec:feature_matching}). 
Finally, at test time, apart from automatically recovering line  correspondences, the $6$-DoF relative pose ($\bR$ and $\bt$) between two coordinate systems is recovered via a $2$-line minimal solver with RANSAC (\secref{sec:pose_solver}). }
\vspace*{-10pt}
\label{fig:overall_framework}
\end{figure}

Our  \plucker Net  is  trained
end-to-end. The overall framework is illustrated in Figure \ref{fig:overall_framework}. Our contributions are:

\begin{enumerate}
\itemsep0em
\item A simple, straightforward and effective learning-based method to estimate a rigid transformation aligning two line reconstructions in Euclidean space;
\item A deep neural network extracting features from lines, while respecting the line geometry;
\item An original global feature matching network based on the optimal transport theory to find line correspondences;
\item A $2$-line minimal solver with RANSAC to register 3D line reconstructions in Euclidean space;
\item We propose two 3D line registration baselines (iterative closet lines and direct regression), three benchmark datasets build upon \cite{Structured3D,hackel2017isprs,apollo_bib} and show the  state-of-the-art performance of our method. 
\end{enumerate}

\section{Related Work}
For space reasons, we focus on geometric deep learning and aligning line reconstructions, hence omitting a large body of works on 2D line detection and description from images~\cite{von2008lsd,akinlar2011edlines,zhang2013efficient,li2016line,zhou2019end,xue2020holistically}, and 3D line fitting~\cite{hofer2015line3d,lu2019fast,wang2020pie}. Interested readers can refer to these papers for details.

\vspace*{-10pt}
\paragraph{Learning from unordered sets.}
PointNet-based~\cite{qi2017pointnet} networks can handle sparse and unordered 3D points. Though most works focus on classification and segmentation tasks \cite{wang2018dynamic,qi2017pointnet}, geometry problems are ready to be addressed. For example, 3D--3D point cloud registration \cite{aoki2019pointnetlk}, 2D--2D \cite{yi2018learning} or 2D--3D \cite{dang2018eigendecomposition} outlier correspondence rejection, and 2D--3D camera pose estimation \cite{liu2020learning,CampbellAndLiu:ECCV2020}.
Instead of studying 3D points, this paper tackles the problem of learning from lines, specifically, aligning line reconstructions, which has not previously been researched. 
\vspace*{-10pt}
\paragraph{Sparse feature matching.}
Two sparse feature sets can be matched via nearest neighbor search with ratio test  \cite{lowe2004distinctive} or Hungarian algorithm \cite{kuhn1955hungarian}. Recently, researchers~\cite{liu2020learning,CampbellAndLiu:ECCV2020,sarlin2020superglue, simon2020stickypillars} bring the Sinkhorn's algorithm \cite{cuturi2013sinkhorn} to this matching task for its end-to-end training ability. Though \cite{CampbellAndLiu:ECCV2020, sarlin2020superglue, simon2020stickypillars,dang2020learning} show they can match sparse feature sets, their usage only focuses on pairwise feature distance or  matching cost. In contrast, we formulate this matching problem in line with traditional optimal transport \cite{villani2009optimal,cuturi2013sinkhorn,courty2016optimal} inside a probability framework, which aims to align two 1D probability distributions by estimating a 2D probabilistic coupling matrix. We directly regress these two 1D probability distributions using a multilayer perceptron (MLP) based network, and use a variant of Sinkhorn's algorithm \cite{sinkhorn1964relationship,marshall1968scaling} to calculate a 2D probability matrix, describing the matchability of these two sparse line sets.
\vspace*{-10pt}
\paragraph{Relative pose estimation.}

Iff with ground-truth matches, traditional methods \cite{bartoli20013d, bartoli2003motion} can be utilized to estimate a relative pose between two line reconstructions. Bartoli~\etal~\cite{bartoli20013d} present a family of line motion matrix in projective, affine, similarity, and Euclidean space. Minimal $7$ line matches are required to solve the line motion matrix linearly. They \cite{bartoli2003motion} further extend this method, and derives a minimal $5$ line matches method. They also give a minimal $2$ line matches solver for solving the line motion matrix in similarity space. This paper aligns line reconstructions in Euclidean space with unknown line-to-line matches.
% With PointNet \cite{}, many 3D points related topics are studied such as ICP,PnP,etc. 
% however, this is not the case for 3D line, which is black at this moment
% it is non-trivial to learning from lines using a deep neural network, as we need to carefully handle line parameterization and geometry in line space.
% This paper studies the problem of learning to register two 3D line reconstructions, and the recoved 6-DoF motion can be used in many areas of computer vision \cite{smith2006real,andreff2002visual,liu1990determination}
% our contributions are : 1) A first, simple, straightforward, yet effective method to predict a rigid transformation aligning two line reconstructions; 2) a deep neural network extracting features from lines, while respecting the knowledge of traditional line-geometry; 3) an original global feature matching network based on a recurrent Sinkhorn layer to find line correspondences; 4) a 2-line minimal solver to register 3D line reconstruction in Euclidean space in a RANSAC framework; 5) propose two 3D line registration baselines, and show the effectiveness of our method. We  will release  our  code,  to  facilitate  reproducibility  and future research.

\section{\plucker Net}
In this section we present \plucker Net -- our method for aligning 3D line reconstructions. We first  define the problem in \secref{sec:ProblemDefinition}, then describe our pipeline. Specifically, We present our method for extracting line-wise discriminative features in \secref{sec:feature_extraction}. We then describe our global feature matching method for obtaining 3D--3D line match probabilities in \secref{sec:feature_matching}. Finally, we provide the relative pose solver, a $2$-line minimal solver in a RANSAC framework in \secref{sec:pose_solver}.

\subsection{Problem Definition}\label{sec:ProblemDefinition}

\paragraph{\plucker \ line.}
A line $\bel$ in a 3D space has $4$ degrees of freedom. Usually, we have three ways to denote a 3D line: 1) a direction and a point the line passes through; 2) two points the line passes through \footnote{Extracting endpoints of lines accurately and reliably is difficult due to viewpoint changes and occlusions \cite{zhang2015building,li2016line}.}, \ie line segment with two start and end junction points; 3) the \plucker \       coordinates  $(\bv,\bmo)$ \cite{hodge1994methods}, where $\bv$ is the $3$-dim direction vector and  $\bmo$ is the $3$-dim moment vector. We choose \plucker \       coordinates to parameterize a 3D line for its uniqueness once fixing its homogeneous freedom, and its mathematical completeness. We fix homogeneous freedom  of a $6$-dim \plucker \ line by first $L_2$ normalizing its direction vector $\bv$ to a unit-sphere, and then set the value of the first dimension of $\bv$ to be greater than $0$. This ensures $\bv$ lie on a hemisphere.

\paragraph{Partial-to-Partial registration.}

Let $L_\mathcal{S} = \{\bel_i\}, i = 1,...,M$ denote  a  3D  line  set  with $M$ \plucker \ lines in  the source frame, $L_\mathcal{T} = \{\bel_j^{'}\}, j=1,...,N$ denote  a  3D  line  set  with $N$ \plucker \ lines in  the target frame, and $\bC \in \bbR^{M \times N}$ denote the correspondence matrix between $L_\mathcal{S}$ and $L_\mathcal{T}$.

We aim to estimate a rotation matrix $\bR$ and a translation vector $\bt$ which transforms source line set $L_\mathcal{S}$ to align with the target line set $L_\mathcal{T}$. Specifically, $\bel_j^{'} \approx \bT \bel_i$ for $\bC_{ij} = 1$, where the line motion matrix $\bT$ \cite{bartoli20013d} is given by: 
\begin{equation} \label{Eq::line_motion_matrix}
    \bT = \begin{pmatrix}
\bR & [\bt]_\times \bR\\ 
0 & \bR
\end{pmatrix},
\end{equation}
where $[\bt]_\times$ is a skew-symmetric matrix.

The difficulty of this registration problem is to estimate the correspondence matrix $\bC$. We propose to estimate $\bC$ using a deep neural network. Specifically, for each tentative line match in $\bC$, we calculate a weight $\bW_{ij}$ describing the matchability of $\bel_i$ and $\bel_j^{'}$. We can obtain a set of line matches by taking the Top-$K$ matches according to these weights. With line matches, we give a minimal  $2$-line method with RANSAC
% extend \cite{bartoli2003motion} in Euclidean space with RANSAC 
to solve the problem.

% Approach:

\subsection{Feature extraction} \label{sec:feature_extraction}

\paragraph{Basic block.} Our input is a set of unordered lines. Inpired by PointNet \cite{qi2017pointnet}  which consumes a set of unordered points, we also use MLP blocks to extract line-wise features. An MLP block, \eg, MLP($128,128$) denotes a two layers perceptron, with each layer size being $128$. In our network,  Groupnorm \cite{wu2018group} with GeLU \cite{hendrycks2016gaussian} is used for all layers (except the last layer) in an MLP block. We found Groupnorm \cite{wu2018group} and GeLU \cite{hendrycks2016gaussian} is better than the commonly used Batchnorm \cite{ioffe2015batch} and ReLU, respectively.

\vspace*{6pt}
\noindent{\textbf{Subspace coding.}}
For a $6$-dim \plucker \ line, its direction $\bv$ and moment $\bmo$ lies in two domains. $\bv$ lies on a hemisphere and $\bmo$ is unbounded with constraint $\bv \cdot \bmo = 0$. $\left \| \bmo \right \|_2$ gives the distance from the origin to the line (iff $\bv$ is $L_2$-normalized). To bridge this domain gap, we first use two parallel networks without sharing weights to process $\bv$ and $\bmo$ independently, and then concatenate features from them to embed a \plucker \ line to a high-dim space \footnote{For features from the two subspace, we find no additional benefit of imposing the orthogonal constraint via the Gram-Schmidt process, though the original direction $\bv$ is orthogonal to the moment $\mathbf{m}$.  }.
 The benefits of this subspace coding process lie in two parts: 1) the domain gap between $\bv$ and $\bmo$ is explicitly considered; 2) we are able to define geometric nearest neighbors in each subspace of direction $\bv$ and moment $\bmo$. Angular distance and $L_2$ distance is used to find geometric nearest neighbors in the subspace of $\bv$ and $\bmo$, respectively. Note that we cannot define nearest neighbors in $6$-dim space, as there is no universally agreed global distance metric between two \plucker \  lines, see discussions in \secref{sec:pose_solver}.

For each $\mathbf{o}_i$ ($\mathbf{o}_i$ can be $\mathbf{v}_i$ or $\mathbf{m}_i$, and $i$ is a line index) in a subspace, we first perform nearest neighbor search and build a line-wise Knn graph. For a Knn graph around the anchor $\mathbf{o}_i$, the edges from $\mathbf{o}_i$ to its neighbors capture the local geometric structure. Similar to EdgeConv \cite{wang2018dynamic}, we concatenate the anchor and edge features, and then pass them through an MLP layer to extract local features around the $i$-th line. Specifically, the operation is defined by
\begin{equation}\label{eq::local_knn}
    E\left(\mathbf{o}_{i}\right) = \text{avg}_{\mathbf{o}_{k}, \mathbf{o}_{k}\in\mho(\mathbf{o}_i)} \left( \btheta \left( \mathbf{o}_{k}-\mathbf{o}_{i} \right) + \bphi \mathbf{o}_{i} \right),
\end{equation}
where $\mathbf{o}_{k}\in\mho(\mathbf{o}_i)$ denotes that $\mathbf{o}_{k}$ is in the neighborhood $\mho(\mathbf{o}_i)$ of $\mathbf{o}_i$, $\btheta$ and $\bphi$ are MLP weights performed on the edge $\left ( \mathbf{o}_{k}-\mathbf{o}_{i} \right )$ and anchor $\mathbf{o}_{i}$ respectively, and $\text{avg}(\cdot)$ denotes that we perform average pooling in the neighborhood $\mho(\mathbf{o}_i)$ after the MLP to extract a single feature for the $i$-th line.

\begin{figure}
\begin{center}
\includegraphics[width=0.48\textwidth]{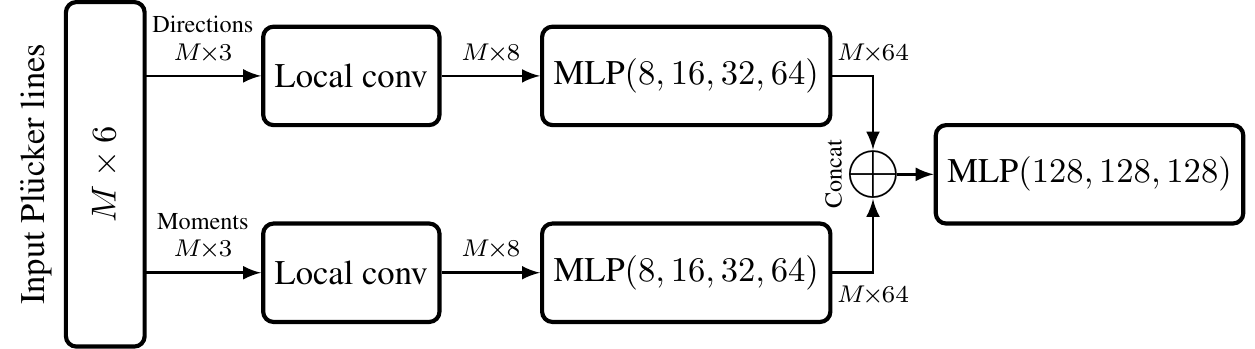}
\end{center}
% \vspace{-10pt}
\caption{\it Pipeline of subspace coding. The network takes
$M$ $6$-dim \plucker \ lines as input, split lines into $M$ $3$-dim moments and directions, extract features in the subspace of moments and directions independently, and finally concatenate features from subspaces to obtain global high-dim \plucker \ line features. Local conv. stands for aggregating line-wise local geometry features and is given in Eq.~\eqref{eq::local_knn}. $\text{MLP}(\cdot)$ stands for multi-layer perceptron block, numbers
in bracket are layer sizes. }
% \vspace{10pt}
\label{fig:subspaceLifting}
\end{figure}

After extracting line-wise local features $E\left(\mathbf{v}_{i}\right)$ and $E\left(\mathbf{m}_{i}\right)$ in the subspace of $\bv$ and $\bmo$ independently, we lift them to high-dim spaces by using an MLP block $\text{MLP}(8,16,32,64)$. Output features from the two subspaces are concatenated, and further processed by an MLP block $\text{MLP}(128,128,128)$ to embed to a $128$-dim space. 
This subspace coding process is given in Figure \ref{fig:subspaceLifting}.

\begin{figure}
\begin{center}
\includegraphics[width=0.48\textwidth]{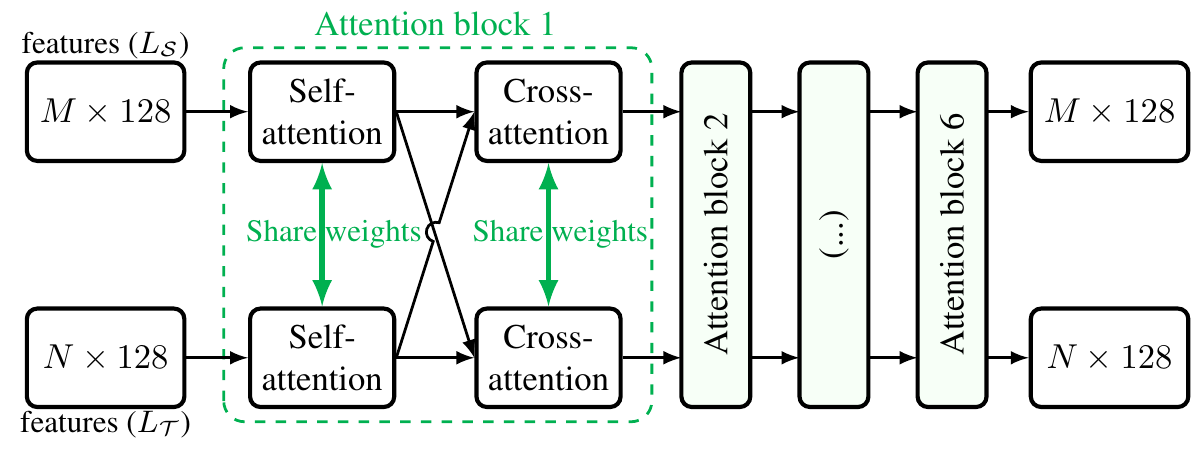}
\end{center}
% \vspace{-10pt}
\caption{\it Pipeline of discriminative feature embedding. The network takes
$M$ and $N$ $128$-dim line features from subspace coding as input, improve feature discriminativeness via both self and cross attention. We use $6$ attention blocks in total. }
\vspace*{-10pt}
\label{fig:discriminativefeatureembedding}
\end{figure}

\vspace*{6pt}
\noindent{\textbf{Discriminative feature embedding.}}
% \paragraph{Discriminative feature embedding}
After coding $6$-dim  \plucker \ lines to  $128$-dim features, we aim to make these features discriminative for matching.
% Note we are solving a two-view feature matching problem.
Inspired by the success of attention mechanism and graph neural network in 3D--3D points cloud registration \cite{Wang_2019_ICCV,dang2020learning,simon2020stickypillars} and two-view image matching \cite{sarlin2020superglue}, we adopt both self-attention and cross-attention to extract line-wise discriminative features. This process is given in Figure \ref{fig:discriminativefeatureembedding}.

We first define two complete graphs for source and target line reconstructions, denoted by $\mathcal{G}_{{L}_\mathcal{S}}$ and $\mathcal{G}_{{L}_\mathcal{T}}$, respectively. Nodes in $\mathcal{G}_{{L}_\mathcal{S}}$ and $\mathcal{G}_{{L}_\mathcal{T}}$ corresponds to lines $\bel_i$ and $\bel_j^{'}$, respectively. Node values corresponds to line features, and are denoted by $^{(t)}\mathbf{f}_{\bel_i}$ and $^{(t)}\mathbf{f}_{\bel_j^{'}}$, where $t$ is the layer index. Initial node values ($t=1$) are outputs of subspace coding.

The two complete graphs ($\mathcal{G}_{{L}_\mathcal{S}}$, $\mathcal{G}_{{L}_\mathcal{T}}$) form a multiplex
graph \cite{mucha2010community, nicosia2013growing}. It contains both intra-frame (or self) edges $\mathcal{E}_{\text{self}}$ and inter-frame (or cross) edges $\mathcal{E}_{\text{cross}}$. Intra-frame edges  connecting all lines within the same line reconstructions.  Inter-frame edges connecting one line from $\mathcal{G}_{{L}_\mathcal{S}}$ to all lines in $\mathcal{G}_{{L}_\mathcal{T}}$, and vice versa.

Node values are updated using multi-head self and cross attention in a message passing framework \cite{sarlin2020superglue, simon2020stickypillars}. The merit of using this framework is integrating both intra-frame and inter-frame contextual cues to increase distinctiveness of line features. For self-contain purposes, we briefly summarize it.

Node values in $\mathcal{G}_{{L}_\mathcal{S}}$ are updated via (same for $\mathcal{G}_{{L}_\mathcal{T}}$):

\begin{equation}\label{Eq::GNN_updated}
    ^{(t+1)}\mathbf{f}_{\bel_i} = ^{(t)}\mathbf{f}_{\bel_i} + \mathcal{U}\left(^{(t)}\mathbf{f}_{\bel_i} || \mathfrak{m}_{\mathcal{E}\to {\bel_i}} \right),
\end{equation}
where $\cdot||\cdot$ denotes concatenation, $\mathfrak{m}_{\mathcal{E}\to \bel_i}$ denotes message from $\mathcal{E}$ to node ${\bel_i}$, and $\mathcal{U}(\cdot)$ denotes feature propagation and is implemented as an MLP block MLP($256,256,128$). The message $\mathfrak{m}_{\mathcal{E}\to \bel_i}$ encodes propagation of all nodes which are connected to node $\bel_i$ via edges $\mathcal{E}$ ($\mathcal{E} = \mathcal{E}_\text{self} \ \text{or} \  \mathcal{E}_\text{cross}$). In our implementation, the total depth of network is set to $12$ (\ie, $t \in [1,T], T=12$) and we alternate to perform self and cross message passing with increasing
network depth, \ie, $\mathcal{E} = \mathcal{E}_\text{self}$ if $t$ is odd and $\mathcal{E} = \mathcal{E}_\text{cross}$ if $t$ is even.

Message $\mathfrak{m}_{\mathcal{E}\to \bel_i}$ is calculated through attention \cite{vaswani2017attention}:

\begin{equation}\label{Eq::attention}
    \mathfrak{m}_{\mathcal{E}\to \bel_i} = \Sigma_{j:(i,j)\in \mathcal{E}} \ \ \  \alpha_{i,j}\mathtt{v}_j,
\end{equation}
where attention weight $\alpha_{i,j}$ is the Softmax over the key $\mathtt{k}_j$ to query $\mathtt{q}_i$ similarites,  $\alpha_{i,j} = \text{Softmax}_j\left( \mathtt{q}_i^{\mathtt{T}}\mathtt{k}_j \right/\sqrt{D})$, $D=128$ is the feature dimension. In our implementation, the key $\mathtt{k}_j$, query $\mathtt{q}_i$ and value  $\mathtt{v}_j$ are obtained via linear projection of line features, using different projection matrix. Following common practice, multi-head ($4$-heads in this paper) attention is used to improve the performance of Eq.~\eqref{Eq::attention}. We do not use recent works \cite{choromanski2020rethinking,wang2020linformer} concerning speed-up the computations of Eq.~\eqref{Eq::attention} as they all decrease the performance of our \plucker Net.

\subsection{Feature matching} \label{sec:feature_matching}
Given a learned feature descriptor per line in ${{L}_\mathcal{S}}$ and ${{L}_\mathcal{T}}$, we perform global feature matching to estimate the likelihood that a given \plucker \ line pair matches. 

\paragraph{Matching cost.}
We first compute the pairwise distance matrix $\bH \in \bbR_{+}^{M\times N}$, which measures the cost of assigning lines in ${{L}_\mathcal{S}}$ to ${{L}_\mathcal{T}}$. To calculate $\bH$, we linearly project outputs $^{(T)}\mathbf{f}_{\bel_i}$ and $^{(T)}\mathbf{f}_{\bel_j^{'}}$ from discriminative feature embedding process to obtain $\mathbf{f}_{\mathbf{x}_i}$ and $\mathbf{f}_{\mathbf{y}_j}$ for lines in ${{L}_\mathcal{S}}$ and ${{L}_\mathcal{T}}$, respectively. $\mathbf{f}_{\mathbf{x}_i}$ and $\mathbf{f}_{\mathbf{y}_j}$ are post $L_2$ normalized to embed them to a metric space.
Each element of $\bH$ is the $L_2$ distance between the features at line $\bel_i$ and $\bel_j^{'}$, \ie, $\bH_{ij} = \| \bbf_{\bx_i} - \bbf_{\by_j} \|_2$.

\paragraph{Prior Matchability.}
We are solving a partial-to-partial registration problem. Lines in ${{L}_\mathcal{S}}$ and ${{L}_\mathcal{T}}$ do not necessarily have to match. 
To model the likelihood that a given line has a match and is not an outlier, we define unary matchability vectors, denoted by $\br$ and $\bs$ for lines in ${{L}_\mathcal{S}}$ and ${{L}_\mathcal{T}}$, respectively. To estimate $\br$ (same for $\bs$), we add a lightweight matchability regression network, and the operation is defined by:

\begin{equation}\label{Eq::unary_prob_est}
\footnotesize
    \br_{\bel_i} = \text{Softmax}_i\left( \mathcal{P}\left( ^{(T)}\mathbf{f}_{\bel_i} || {\text{avg}_{\bel_j^{'}\in L_\mathcal{T}}} ^{(T)}\mathbf{f}_{\bel_j^{'}} || {\text{max}_{\bel_j^{'}\in L_\mathcal{T}}} ^{(T)}\mathbf{f}_{\bel_j^{'}} \right) \right),
\end{equation}
where ${\text{avg}_{\bel_j^{'}\in L_\mathcal{T}}} (\cdot)$ and ${\text{max}_{\bel_j^{'}\in L_\mathcal{T}}} (\cdot)$ denotes performing average and max pooling at each feature dimension, aiming to capture global context of line features for $L_\mathcal{T}$. $\mathcal{P}(\cdot)$ denotes feature propogation and is implemented as an MLP block MLP$(384,256,256,128,1)$. Softmax is used to convert lines' logits to probabilities. Eq.~\eqref{Eq::unary_prob_est} is based on the idea of looking at global context of cross frame $L_\mathcal{T}$  to regress line-wise matching prior for current frame $L_\mathcal{S}$.

Collecting the matchabilities for all lines in $L_\mathcal{S}$ or $L_\mathcal{T}$ yields a matchability histogram, a 1D probability distribution, given by $\br \in \Sigma_M$ and $\bs \in \Sigma_N$, where a simplex in $\bbR^M$ is defined as
$\Sigma_{M} = \left\{ \br \in \bbR_+^M , \sum_i \br_i = 1 \right\}$.

\begin{figure}
\begin{center}
\includegraphics[width=0.49\textwidth]{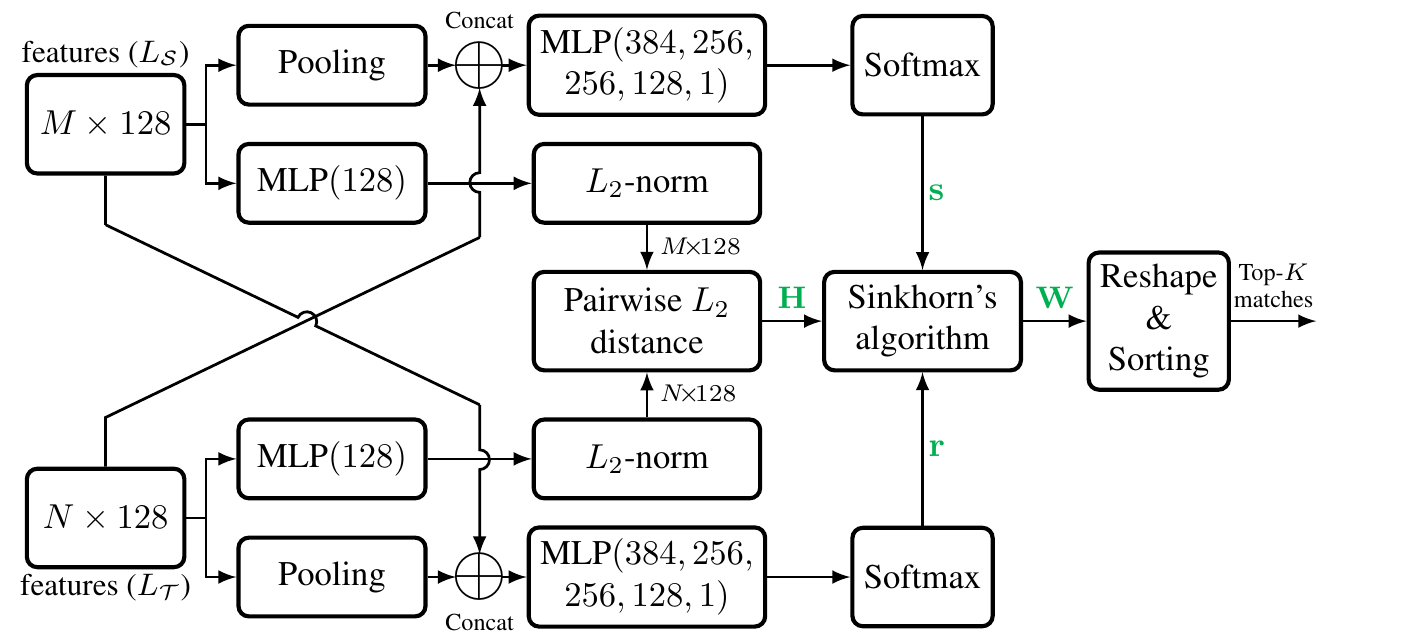}
\end{center}
% \vspace{-10pt}
\caption{\it Our feature matching pipeline. Given an $M \times 128$ feature set from discriminative feature embedding of source line set $L_{\mathcal{S}}$ and an $N \times 128$ feature set from the target line set $L_{\mathcal{T}}$, we compute the pairwise $L_2$ distance matrix $\bH$. Along with a unary matchability $M$-vector $\mathbf{r}$ from $L_{\mathcal{S}}$ and $N$-vector $\mathbf{s}$ from $L_{\mathcal{T}}$, the distance matrix $\bH$ is transformed to a joint probability matrix $\bW$ using Sinkhorn's algorithm. Reshaping $\bW$ and sorting the line matches by their corresponding matching probabilities generates a prioritized line match list. We take the Top-$K$ matches as our set of putative correspondences. }
% \vspace{10pt}
\label{fig:feature_matching}
\end{figure}

\paragraph{Global matching.}
From $\bH$, $\br$, and $\bs$, we estimate a weighting matrix $\bW \in \bbR_{+}^{M\times N}$ where each element $\bW_{ij}$ represents the matchability of the \plucker \ line pair $\{\bel_i, \bel_j^{'}\}$. Note that each element $\bW_{ij}$ is estimated from the cost matrix $\bH$ and the unary matchability vectors $\br$ and $\bs$,  rather than locally from $\bH_{ij}$. In other words, the weighting matrix $\bW$ globally handles pairwise descriptor distance ambiguities in $\bH$, while respecting the unary priors. The overall pipeline is given in Figure \ref{fig:feature_matching}.

\paragraph{Sinkhorn solver.}
From optimal transport theory \cite{villani2009optimal,cuturi2013sinkhorn,courty2016optimal}, the joint probability matrix $\bW$ can be solved by
\begin{equation}\label{eq::OT}
  \argmin_{\bW\in\Pi\left(\br,\bs \right)} \left \langle \bH,\bW \right \rangle - \lambda E\left( \bW \right),
\end{equation}
where $\left\langle\cdot,\cdot\right\rangle$ is the Frobenius dot product and $\Pi(\br,\bs)$ is the transport polytope that couples two unary matchability vectors $\br$ and $\bs$, given by
\begin{equation}\label{eq::Pi}
\Pi\left(\br,\bs \right) = \left \{\bW \in  \bbR_+^{M\times N}, \bW\mathbf{1}^N = \br, \bW\transpose\mathbf{1}^M = \bs \right \},
\end{equation}
where $\mathbf{1}^N = [1,1,...,1]\transpose\in \bbR^N$.
The constraint on $\bW$ ensures that we assign the matchabilities of each line in $L_\mathcal{S}$ (or $L_\mathcal{T}$) to all lines in $L_\mathcal{T}$ (or $L_\mathcal{S}$) without altering its matchability.
The entropy regularization term $E\left(\bW \right)$
%encourages $\bW$ to have a uniform distribution
facilitates efficient computation \cite{cuturi2013sinkhorn} and is defined by
\begin{equation}
    E\left (\bW \right ) = -\sum_{i,j}{\bW}_{ij}\left ( \log \bW_{ij} - 1\right ).
\end{equation}
% Note that when $\lambda = 0$, equation \eqref{eq::OT} is the classical (discrete) optimal transport \cite{villani2009optimal}.  

To solve \eqnref{eq::OT}, we use a variant of Sinkhorn's Algorithm \cite{sinkhorn1964relationship,marshall1968scaling}, given in \algoref{alg:sinkhorn}.
Unlike the standard algorithm that generates a square, doubly-stochastic matrix, our usage generates a rectangular joint probability matrix, whose existence is guaranteed (Theorem $4$ \cite{marshall1968scaling}).

\begin{algorithm}
\SetAlgoLined
\DontPrintSemicolon
\KwData{$\bH$, $\br$, $\bs$, $\bb = \textbf{1}^N$, $\lambda$, and iterations $\textit{Iter}$}
\KwResult{Weighting matrix $\bW$}
% \textbf{Input:} $\bH$, $\br$, $\bs$, $\bb = \textbf{1}^N$, $\lambda$, and iterations $\textit{Iter}$\\
$\bm{\Upsilon} = \exp\left (-\textbf{H}/\lambda  \right ) $ \\
$\bm{\Upsilon} = \bm{\Upsilon} / \sum\bm{\Upsilon}$\\
% \tcp*{normalize $\bm{\Upsilon}$ to be a joint probability matrix}
\While{$\text{it} < \text{Iter}$}{ 
  $\ba = \br \oslash (\bm{\Upsilon}\bb) $;
%   \ \ \ \ \ \ \ \  \Comment{alternatively updating $\ba$ and $\bb$ }
% \tcp*{alternatively updating $\ba$ and $\bb$}
  $\bb = \bs \oslash (\bm{\Upsilon}\transpose\ba) $
 }
$\bW = \diag(\ba)\bm{\Upsilon}\diag(\bb)$
% \tcp*{assemble to build the weighting matrix $\bW$}
\caption{Sinkhorn's Algorithm to solve \eqref{eq::OT}. Hadamard (elementwise) division is denoted by $\oslash$. }\label{alg:sinkhorn}
\end{algorithm}

\paragraph{Loss.} To train our feature extraction and matching network, we apply a loss function to the weighting matrix $\bW$.
Since $\bW$ models the joint probability distribution of $\br$ and $\bs$, we can maximize the joint probability of inlier correspondences and minimize the joint probability of outlier correspondences using

\begin{equation}\label{eq::loss}
    \mathcal{L} = \sum_{i}^{M}\sum_{j}^{N} \gamma_{i,j} \mathcal{F}\left( \bC_{ij}^\text{gt}, {\bW}_{ij} \right),
\end{equation}
where $\bC_{ij}^\text{gt}$ equals to $1$ if $\{\bel_i, \bel_j^{'}\}$ is a true correspondence and $0$ otherwise. $\mathcal{F}\left( \bC_{ij}^\text{gt}, {\bW}_{ij} \right)$ and $\gamma_{i,j}$ are given by,
\begin{equation}
\footnotesize
  \begin{cases}
\mathcal{F}\left( \bC_{ij}^\text{gt}, {\bW}_{ij} \right)=-\log {\bW}_{ij} ; \gamma_{i,j} = 1/\mathcal{A}_\text{true} & \text{ if } \bC_{ij}^\text{gt}=1  \\
\mathcal{F}\left( \bC_{ij}^\text{gt}, {\bW}_{ij} \right)=-\log \left( 1- {\bW}_{ij} \right); \gamma_{i,j} = 1/\mathcal{A}_\text{false} & \text{ if } \bC_{ij}^\text{gt}=0,  \\
\end{cases}  
\end{equation}
where $\gamma_{i,j}$ is a weight to balance true and false correspondences. $\mathcal{A}_\text{true}$ and $\mathcal{A}_\text{false}$ are the total number of true and false correspondences, respectively.

% $\mathcal{F}\left( \bC_{ij}^\text{gt}, {\bW}_{ij} \right) =  -\log {\bW}_{ij}$ for $\bC_{ij}^\text{gt}=1$, and  $\mathcal{F}\left( \bC_{ij}^\text{gt}, {\bW}_{ij} \right) =  -\log \left( 1- {\bW}_{ij} \right)$ for $\bC_{ij}^\text{gt}=0$. $\gamma_{i,j}$ is a weight to balance true and false correspondences. $\gamma_{i,j}$ is set to $1/\mathcal{A}_\text{true}$ for $\bC_{ij}^\text{gt}=1$ and $1/\mathcal{A}_\text{false}$ for $\bC_{ij}^\text{gt}=0$, where ${A}_\text{true}$
% This loss function treats all correspondences as equally important
% We also treat all matchable feature pairs as equally important since the summation of their probabilities are maximized.

% If ground-truth correspondence labels $\bC_{ij}^\text{gt}$ are not available, they can be obtained in a weakly-supervised fashion by transforming lines from  $L_\mathcal{S}$ to $L_\mathcal{T}$ using the ground-truth relative pose and applying an nearest neighbor search to build $\bC_{ij}^\text{gt}$.

\subsection{Pose estimation} \label{sec:pose_solver}
We now have a set of putative line correspondences, some of which are outliers, and want to estimate the relative pose between source and target line reconstructions. According to \cite{bartoli2003motion}, we need minimal $2$ line correspondences to solve the relative rotation $\bR$ and translation $\bt$ in similarity space. In Euclidean space, by substituting Eq.~\eqref{Eq::line_motion_matrix} to $\bel_j^{'} = \bT \bel_i$, we obtain,
\begin{align}
    \bmo_j^{'} &= \bR\bmo_i + [\bt]_\times \bR \bv_i \label{eq::align_trans} \\
    \bv_j^{'} &= \bR\bv_i.  \label{eq::align_direction}
\end{align}
Note $\bt$ is not contained in Eq.~\eqref{eq::align_direction}. We can first solve $\bR$, then substitute $\bR$ to Eq.~\eqref{eq::align_trans} to solve $\bt$.

According to Eq.~\eqref{eq::align_direction}, $\bR$ aligns line direction $\bv_i$ to $\bv_j^{'}$. In \cite{horn1988closed}, authors show that $\bR$ is the closest orthonormal matrix to $\bM = \sum_{}^{} \bv_j^{'}\bv_i^{\transpose}$.  Let $\bM = \bU \bSigma \bV^{\transpose}$ be the singular value decomposition of $\bM$, then $\bR = \bU \bV^{\transpose}$. The sign ambiguity of $\bR$ is fixed by $\bR = \bR / \det(\bR)$.

Substitute estimated $\bR$ to Eq.~\eqref{eq::align_trans}, and reshape it as:
\begin{equation}
     [\bR \bv_i]_\times^{\transpose} \bt = \bmo_j^{'} - \bR\bmo_i.
\end{equation}
Stacking along rows for $i=1,2$ yields a linear equation $\bA\bt = \bb$, where $\bA$ is a $6\times3$ matrix and $\bb$ is a $6\times1$ vector. The least square solution is given by $\bt = \bA^{+}\bb$, where $\bA^{+}$ is the pseudo-inverse of $\bA$.

\paragraph{$2$-line minimal solver in RANSAC framework.} Given the above $2$-line minimal solver, we are ready to integrate it into the RANSAC framework. We need to define the score function, which is used to evaluate the number of inlier line correspondences given an estimated pose from the minimal solver. Our score function is defined by:
\begin{equation}\label{Eq::l2_matric}
    \sS(\bel^{'}, \bel) =  \left \| \bel^{'} - \bT\bel \right \|_{2},
\end{equation}
where we use estimated line motion matrix $\bT$ (Eq.~\eqref{Eq::line_motion_matrix}) to transform $6$-dim \plucker \ lines from $L_\mathcal{S}$ to $L_\mathcal{T}$, and measure the $L_2$ distance between $6$-dim \plucker \ lines.

We choose to use $L_2$ distance for its simplicity, and quadratic for easy-minimization. Though $L_2$ distance only works in a neighborhood of $\bel^{'}$ ($\bv$ and $\bmo$ lie in two different subspaces), in practice, transformed lines from $L_\mathcal{S}$ using our estimated pose lie within the local neighborhood of their matching line in   $L_\mathcal{T}$ (if have), leading to the success of using $L_2$ distance. Despite \plucker \ lines lie in a $4$-dim Klein manifold \cite{pottmann2009computational}, we do not use distance defined via Klein quadric as it needs to specify two additional planes for each line pair, which introduces many hyper-parameters~\cite{pottmann2004line}.

For a \plucker \ line matching pair, if the score function defined in Eq.~\eqref{Eq::l2_matric} is smaller than a  pre-defined threshold $\epsilon$, it is deemed as an inlier pair. After RANSAC, all inlier matching pairs are used to jointly optimize Eq.~\eqref{Eq::l2_matric}.

\section{Experiments}
\subsection{Datasets and evaluation methodology}
We first conduct experiments on both indoor (Structured3D \cite{Structured3D}) and outdoor (Semantic3D \cite{hackel2017isprs}) datasets, and then show our results of addressing real-world line-based visual odometry on the Apollo dataset \cite{apollo_bib}. Sample 3D line reconstructions from these datasets are given in the appendix.

\paragraph{Structured3D~\cite{Structured3D}}

contains 3D annotations of junctions and lines for indoor houses. It has $3,500$ scenes/houses in total, with average/median number of lines at $306$/$312$, respectively. The average size of a house is around $11\text{m}\times10\text{m}\times3\text{m}$. Since the dataset captures structures of indoor houses, most lines are parallel or perpendicular to each other. We randomly split this dataset to form a training and testing dataset, with numbers at $2975$ and $525$, respectively.

% \paragraph{SceneCity3D \cite{zhou2019learning}} contains 3D wireframes for CAD models of outdoor city buildings. It has $20,490$ scenes/street-views in total, with average/median number of lines at $744$/XXX, respectively.  

\paragraph{Semantic3D~\cite{hackel2017isprs}} contains large-scale and densely-scanned 3D points cloud of diverse urban scenes. We use the semantic-8 dataset, and it contains $30$ scans, with over a billion points in total. For each scan, we use a fast 3D line detection method \cite{lu2019fast} to extract 3D line segments. Large-scale 3D line segments are further partitioned into different geographical scenes/cells, with cell size at $10\text{m}\times10\text{m}$ for the $X$--$Y$ dimensions. Scenes with less than $20$ lines are removed. We obtain $1,981$ scenes in total, with average/median number of lines at $676$/$118$, respectively. We randomly split this dataset to form a training and testing dataset, with numbers at $1683$ and $298$, respectively.

\paragraph{Partial-to-Partial registration.}
For 3D lines of each scene ($L_\sS$), we sample a random rigid transformation along each axis, with rotation in $[0^{\circ},45^{\circ}]$ and translation in $[-2.0\text{m},2.0\text{m}]$, and apply it to the source line set $L_\sS$ to obtain the target line set $L_\sT$. We then add noise to lines in $L_\sS$ and $L_\sT$ independently. Specifically, we first transform \plucker \  representation of a line to point-direction representation, with the point at the footprint of line's perpendicular through the origin. Gaussian noise sampled from $\mathcal{N}(0\text{m},0.05\text{m})$
and clipped to $[-0.25\text{m},0.25\text{m}]$ is added to the footprint point, and the direction is perturbed by a random rotation, with angles sampled form $\mathcal{N}(0^{\circ},2^{\circ})$
and clipped to $[-5^{\circ},5^{\circ}]$. After adding noises, point-direction representations are transformed back to \plucker \  representations. To simulate partial scans of $L_\sS$ and $L_\sT$, we randomly select $70\%$ lines from $L_\sS$ and $L_\sT$ independently, yielding an overlapping ratio at $\sim 0.7$.

\paragraph{Implementation details.} 
Our network is implemented in Pytorch and is trained from scratch using the Adam optimizer \cite{kingma2014adam} with a learning rate of $10^{-3}$ and a batch size of $12$. The number of nearest neighbors for each line is set to $10$ in Eq.~\eqref{eq::local_knn}.
The number of Sinkhorn iterations is set to $30$, $\lambda$ is set to $0.1$, Top-$200$ line matches are used for pose estimation, the inlier threshold $\epsilon$ is set to $0.5\text{m}$, and the number of RANSAC iterations is set to $1000$.
Our model is trained on a single NVIDIA Titan XP GPU in one day.

\paragraph{Evaluation metrics.}
% We report the number of inlier 2D--3D matches among all matches found, using ground-truth correspondence labels.
The rotation error is given by the angle difference $\zeta = \arccos((\trace(\bR_\text{gt}\transpose \bR)-1)/{2})$, where $\bR_\text{gt}$ and $\bR$ is the ground-truth and estimated rotation, respectively. The translation error is given by the $L_2$ distance between the ground-truth and estimated translation vector.
We also calculate the recalls (percentage of poses) by varying pre-defined thresholds on rotation and translation error. For each threshold, we count the number of poses with error less than that threshold, and then normalize the number by the total number of poses. 

\paragraph{Baselines.}
There is no off-the-shelf baseline available. We propose and implement:  
\textbf{1) ICL.}  an Iterative-Closest-Line (ICL) method, mimicking the pipeline of traditional Iterative-Closest-Point (ICP) method.  \textbf{2) Regression. } This baseline does not estimate line-to-line matches. After extracting line-wise features, we append a global max-pooling layer to obtain a global feature for each source and target set. A concatenated global feature is fed to an MLP block to directly regress the rotation and translation. Details are given in the appendix.

\subsection{Results and discussions }

\begin{figure}
\begin{center}
\includegraphics[width=0.23\textwidth]{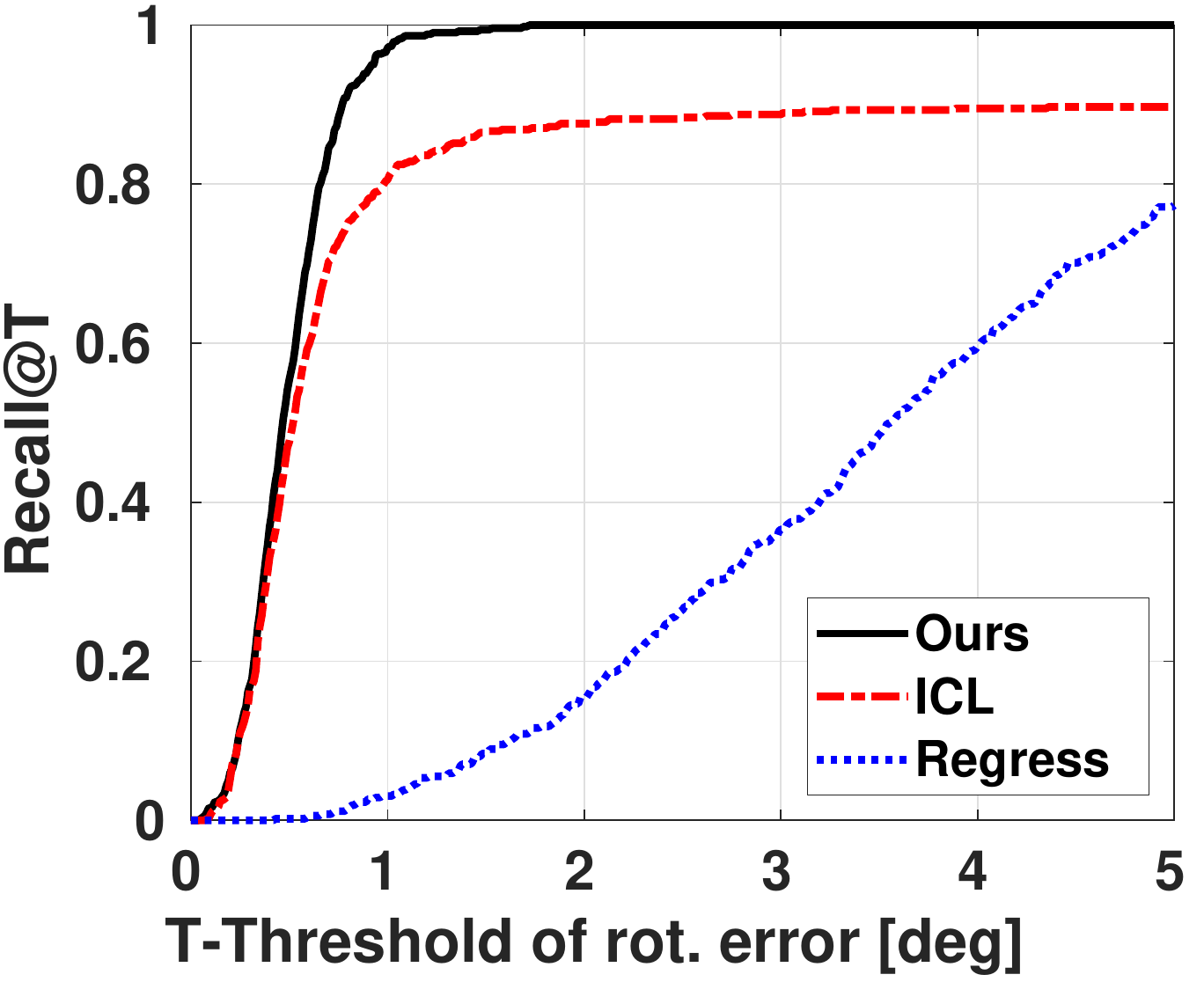}
\includegraphics[width=0.235\textwidth]{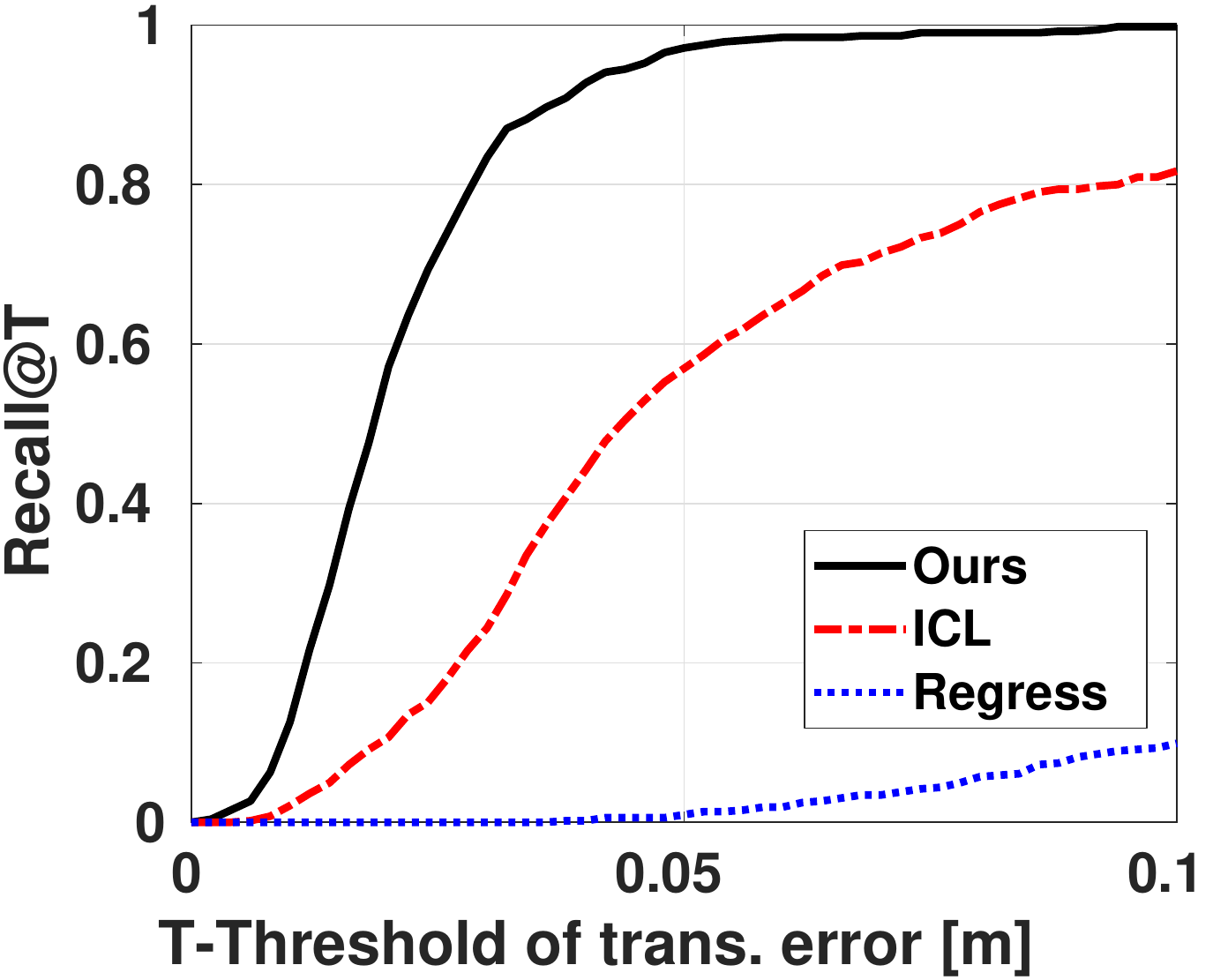}
\includegraphics[width=0.23\textwidth]{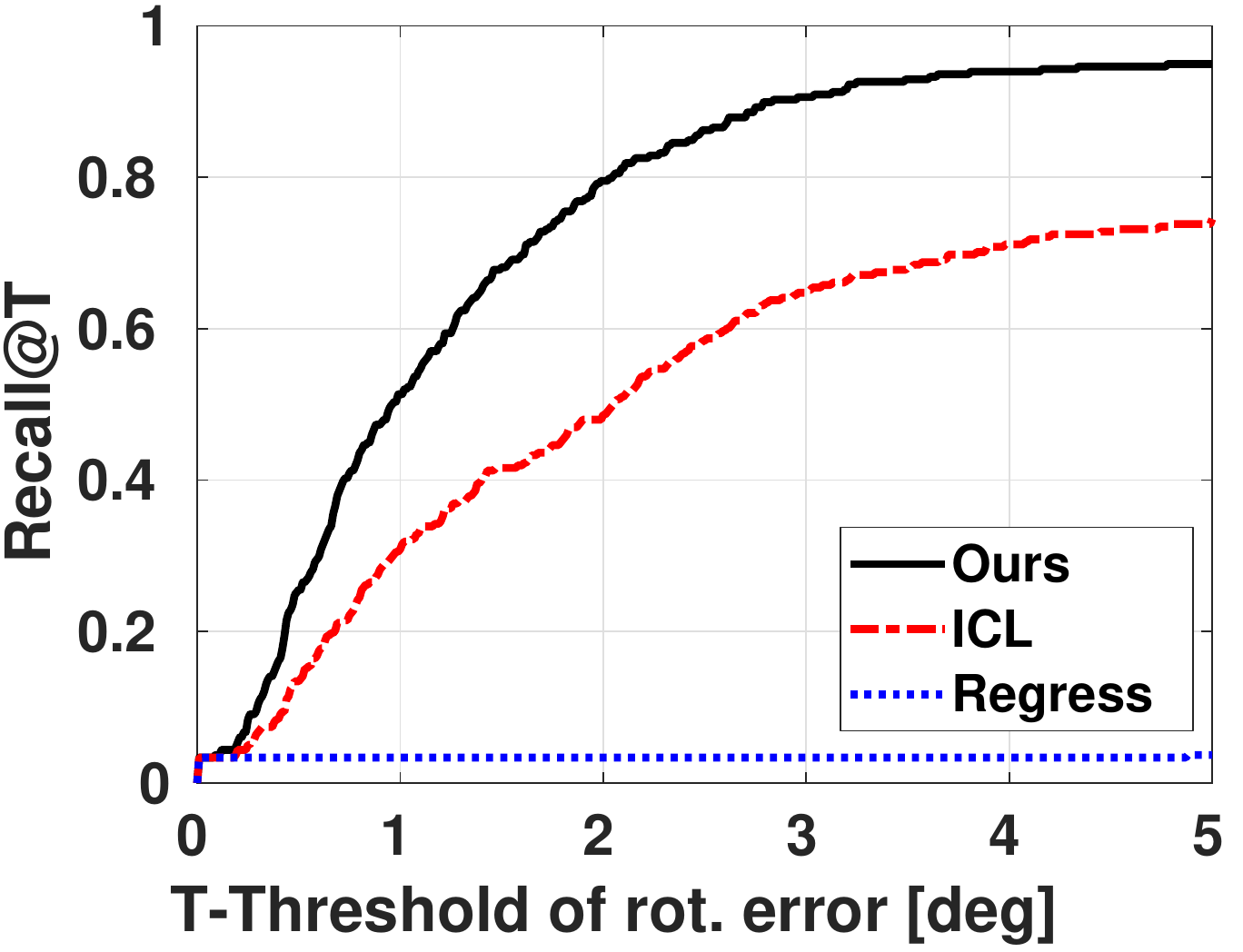}
\includegraphics[width=0.235\textwidth]{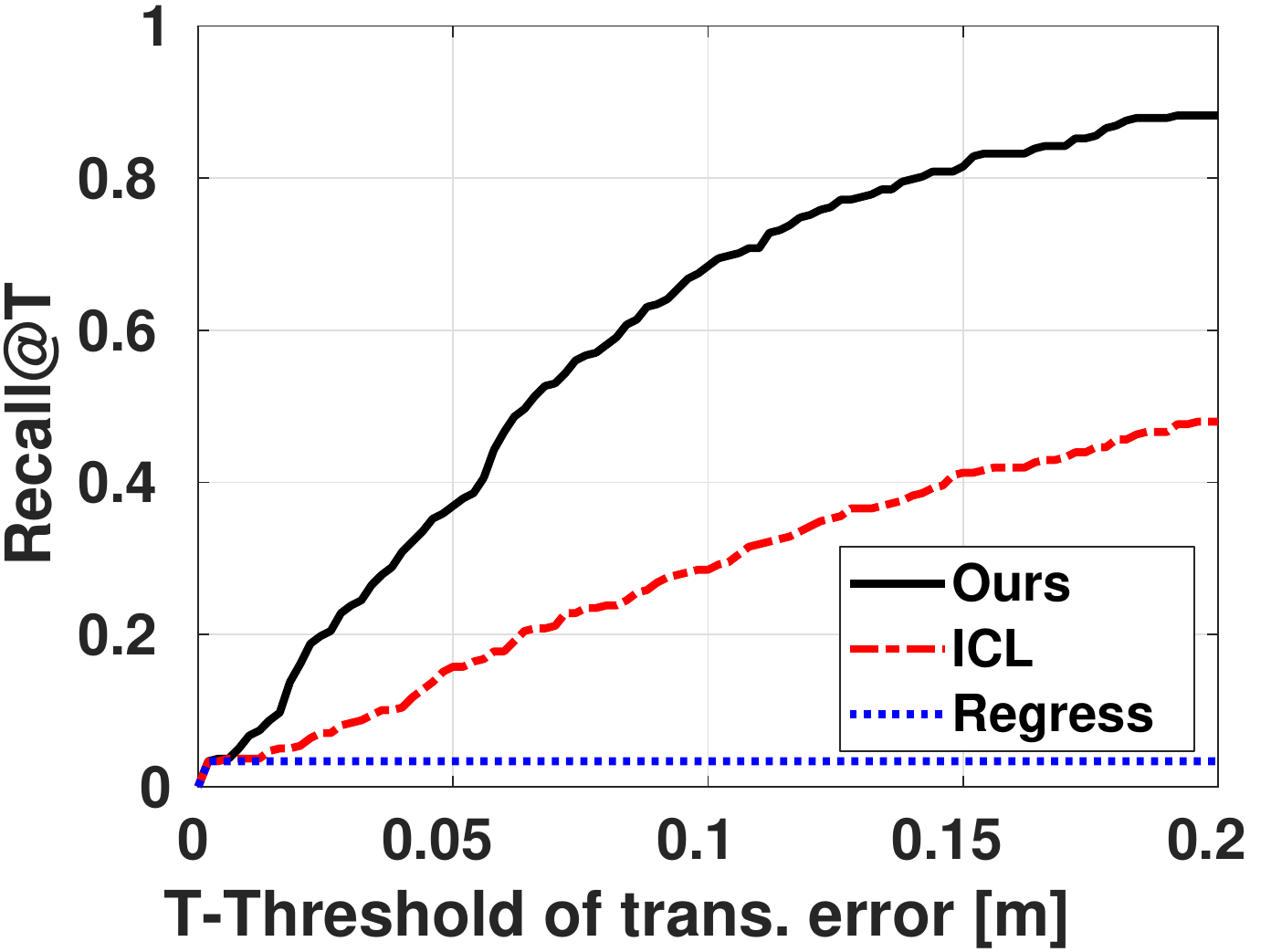}
\end{center}
% \vspace{-10pt}
\caption{\it Recall of rotation (Left) and translation (Right) on  the  Structured3D (Top-row) and Semantic3D (Bottom-row)  datasets,  with respect to an error threshold. The regression baseline diverges on the Semantic3D dataset, leading to the poor performance.}
% \vspace{10pt}
\label{fig:recall_baseline_Structure3Dsemantic3D}
\end{figure}

\paragraph{Comparison with baselines.}
In this experiment, we show the effectiveness of our method to estimate a 6-DoF pose. We compare our method against baselines. The pose estimation performance is given in Table \ref{tab::rotation_trans_error_baseline_Structure3DSemantic3D} and Figure \ref{fig:recall_baseline_Structure3Dsemantic3D}. It shows that our method outperforms baselines with much higher recalls at each level of pre-defined error thresholds. For example, the median rotation and translation error on Semantic3D dataset for our method, ICL and regression is $0.961^{\circ}/2.050^{\circ}/20.934^{\circ}$ and $0.064\text{m}/0.226\text{m}/1.871\text{m}$, respectively. We found that the regression baseline can be trained to converge on the Structured3D dataset, while diverges on the Semantic3D dataset. This shows that directly regressing the relative pose between line reconstructions falls short of applying to different datasets. Since both our method and ICL outperforms the regression baseline significantly, we only compare our method with ICL for  following experiments.

% \begin{figure}
% \begin{center}
% \includegraphics[width=0.23\textwidth]{Figure/semantic3Drot_cmp_baselines.pdf}
% \includegraphics[width=0.235\textwidth]{Figure/semantic3Dtrans_cmp_baselines.pdf}
% \end{center}
% \caption{Recall of rotation (Left) and translation (right) on  the  semantic3D dataset,  with respect to an error threshold.}
% % \vspace{10pt}
% \label{fig:recall_baseline_semantic3D}
% \end{figure}

\begin{table*}[]
 \aboverulesep=0ex
 \belowrulesep=0ex
\setlength{\tabcolsep}{3.8pt}
\small
% \vspace{-4pt}
\caption{\it Comparison of rotation and translation errors on the Structured3D and Semantic3D dataset. Q1 denotes the first quartile, Med. denotes the median, and Q3 denotes the third quartile.}
\vspace*{2pt}
\begin{tabularx}{\linewidth}{llXXXXXX|XXXXXX}
\hline
\toprule
\multicolumn{2}{l}{\multirow{3}{*}{\backslashbox{Method}{Error}}} &
\multicolumn{6}{c|}{Structured3D} &
\multicolumn{6}{c}{Semantic3D}   \\ \cline{3-14} \multicolumn{2}{l}{}                  & 
\multicolumn{3}{c}{Rotation ($^{\circ}$) }  & \multicolumn{3}{c|}{Translation (m)} & 
\multicolumn{3}{c}{Rotation ($^{\circ}$) }  & \multicolumn{3}{c}{Translation (m)} \\ 
\cline{3-14} 
\multicolumn{2}{l}{}                  & Q1    & Med.    & Q3     & Q1     & Med.    & Q3 & Q1    & Med.    & Q3     & Q1     & Med.    & Q3     \\ \hline
\multicolumn{2}{l}{ICL}             & 0.353 & 0.520 & 0.795 & 0.030  & 0.044  & 0.078 & 0.803 & 2.050 & 5.239 & 0.084  & 0.226  & 0.881  \\ 
\multicolumn{2}{l}{Regression}             & 2.436 & 3.610 & 4.935 & 0.151  & 0.240  & 0.367 & 15.902 & 20.934 & 24.947 & 1.347  & 1.871  & 2.281  \\  \hline
\multicolumn{2}{l}{Ours}               & \textbf{0.342} & \textbf{0.468}  & \textbf{0.621}  & \textbf{0.013}  & \textbf{0.019}  & \textbf{0.026} & \textbf{0.482} & \textbf{0.961}  & \textbf{1.791}  & \textbf{0.032}  & \textbf{0.064}  & \textbf{0.119}  \\ \bottomrule
\end{tabularx}
\label{tab::rotation_trans_error_baseline_Structure3DSemantic3D}
\vspace*{-10pt}
\end{table*}

% \begin{table}[]
% \setlength{\tabcolsep}{3.8pt}
% \scriptsize
% % \vspace{-4pt}
% \caption{Comparison of rotation and translation errors on the Semantic3D dataset.}
% \begin{tabularx}{\linewidth}{llXXXXXX}
% \hline
% \multicolumn{2}{l}{\multirow{2}{*}{\backslashbox{Method}{Error}}} & \multicolumn{3}{c}{Rotation ($^{\circ}$) }  & \multicolumn{3}{c}{Translation (m)} \\ 
% \cline{3-8} 
% \multicolumn{2}{l}{}                  & Q1    & Med.    & Q3     & Q1     & Med.    & Q3     \\ \hline
% \multicolumn{2}{l}{ICL}             & 0.803 & 2.050 & 5.239 & 0.084  & 0.226  & 0.881  \\ 
% \multicolumn{2}{l}{Regression}             & 15.902 & 20.934 & 24.947 & 1.347  & 1.871  & 2.281  \\  \hline
% \multicolumn{2}{l}{Ours}               & \textbf{0.482} & \textbf{0.961}  & \textbf{1.791}  & \textbf{0.032}  & \textbf{0.064}  & \textbf{0.119}  \\ \hline
% \end{tabularx}
% \label{tab::rotation_trans_error_baseline_Semantic3D}
% \end{table}

\begin{figure}
\begin{center}
\includegraphics[width=0.23\textwidth]{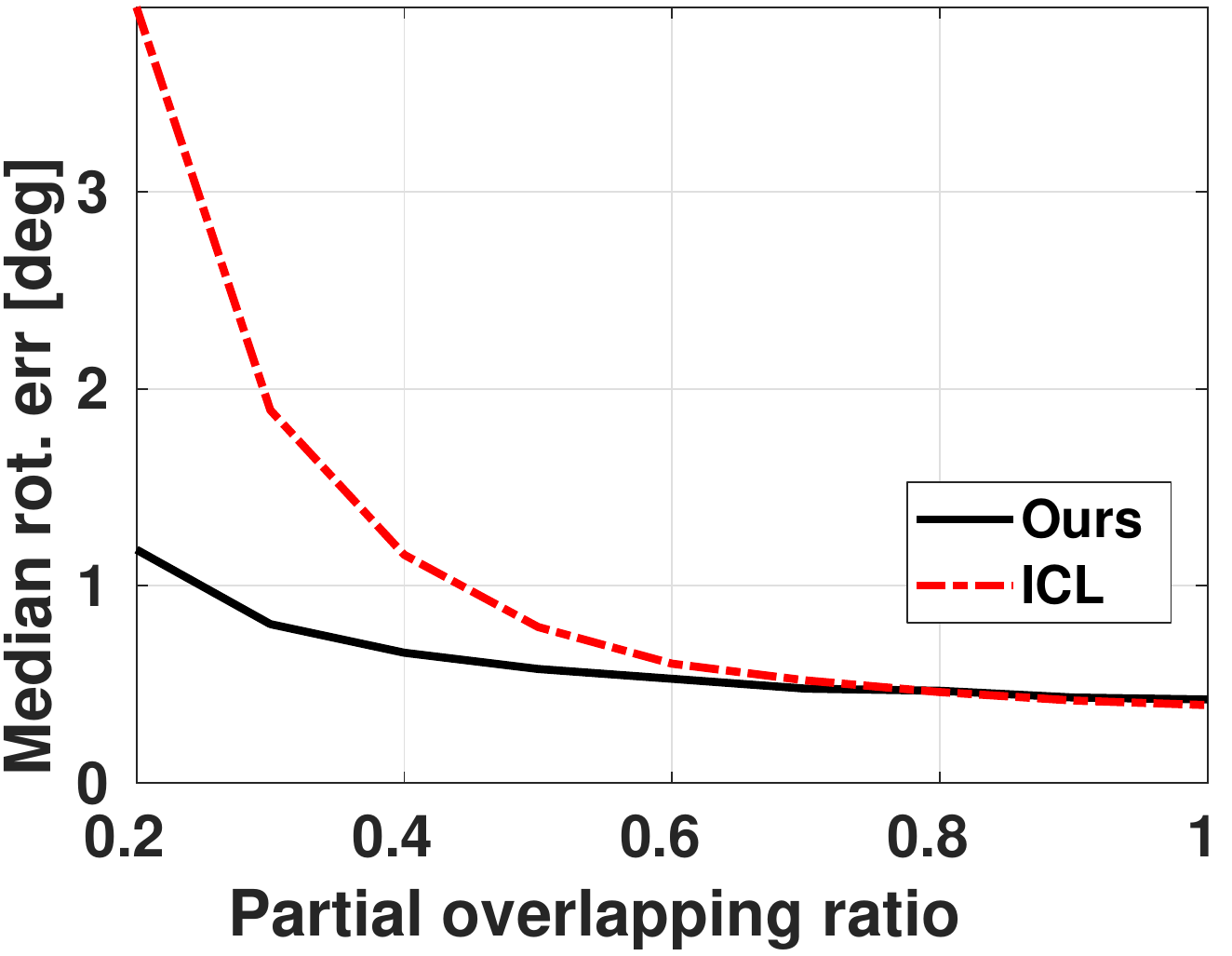}
\includegraphics[width=0.222\textwidth]{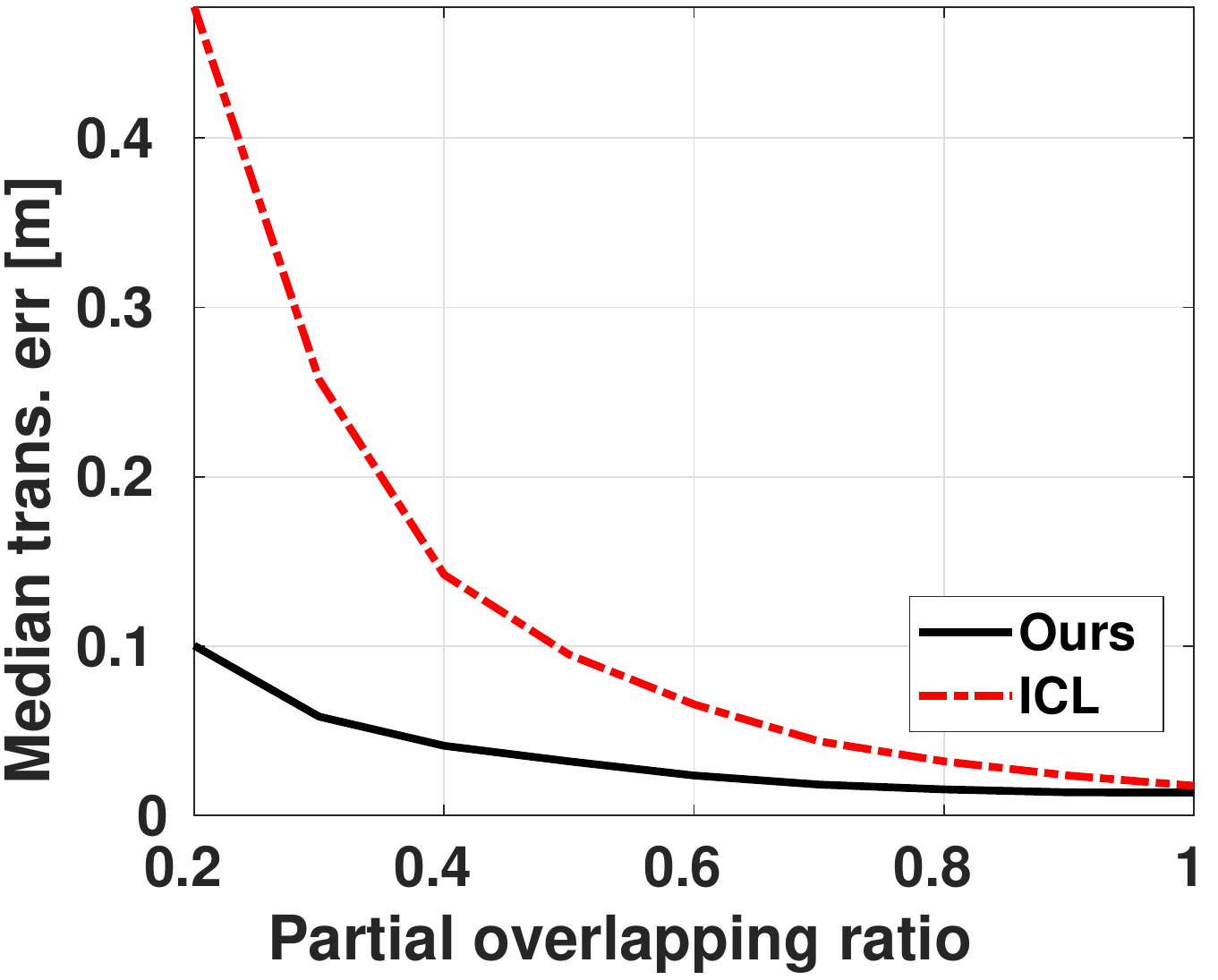}
\includegraphics[width=0.23\textwidth]{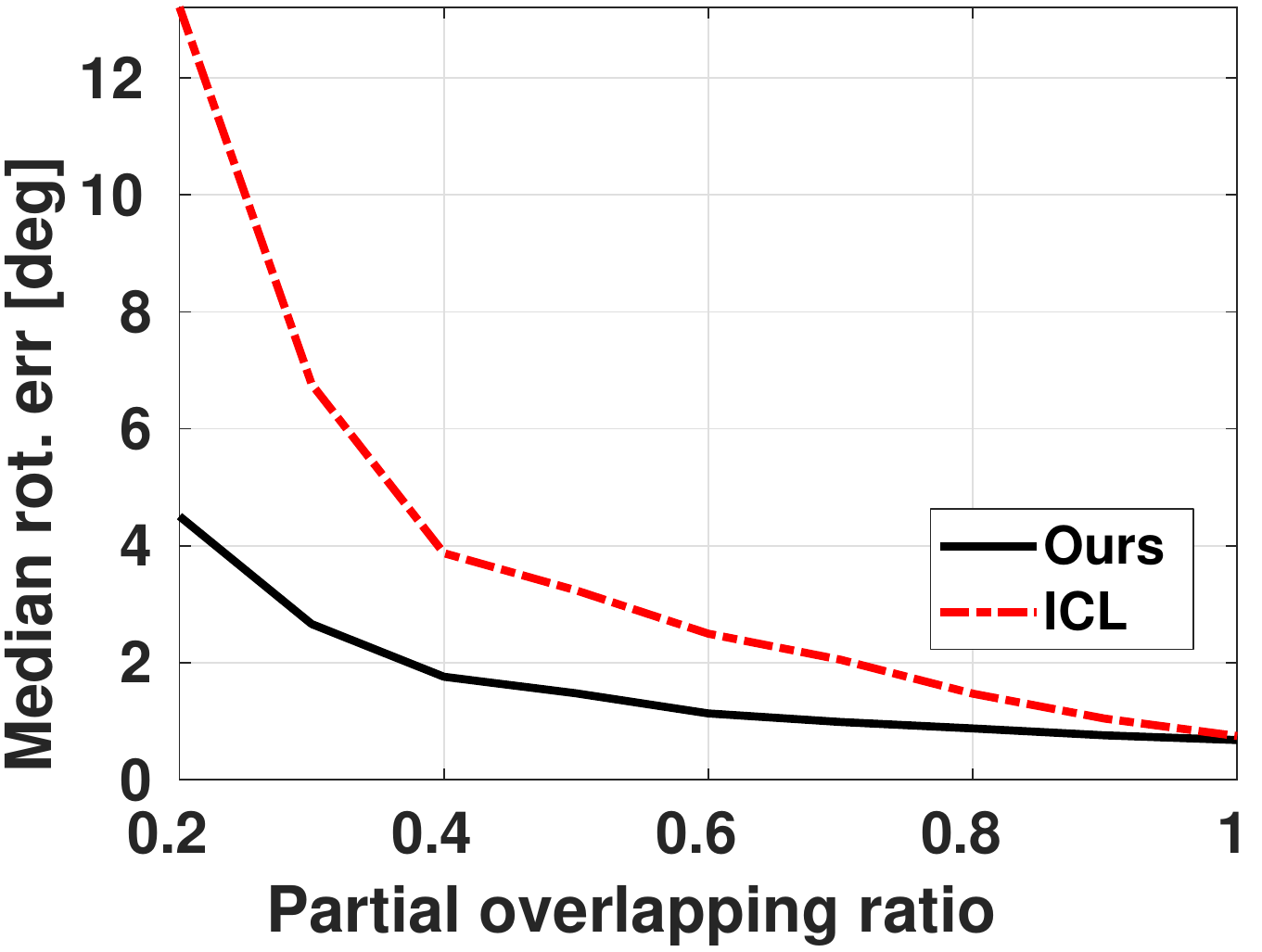}
\includegraphics[width=0.23\textwidth]{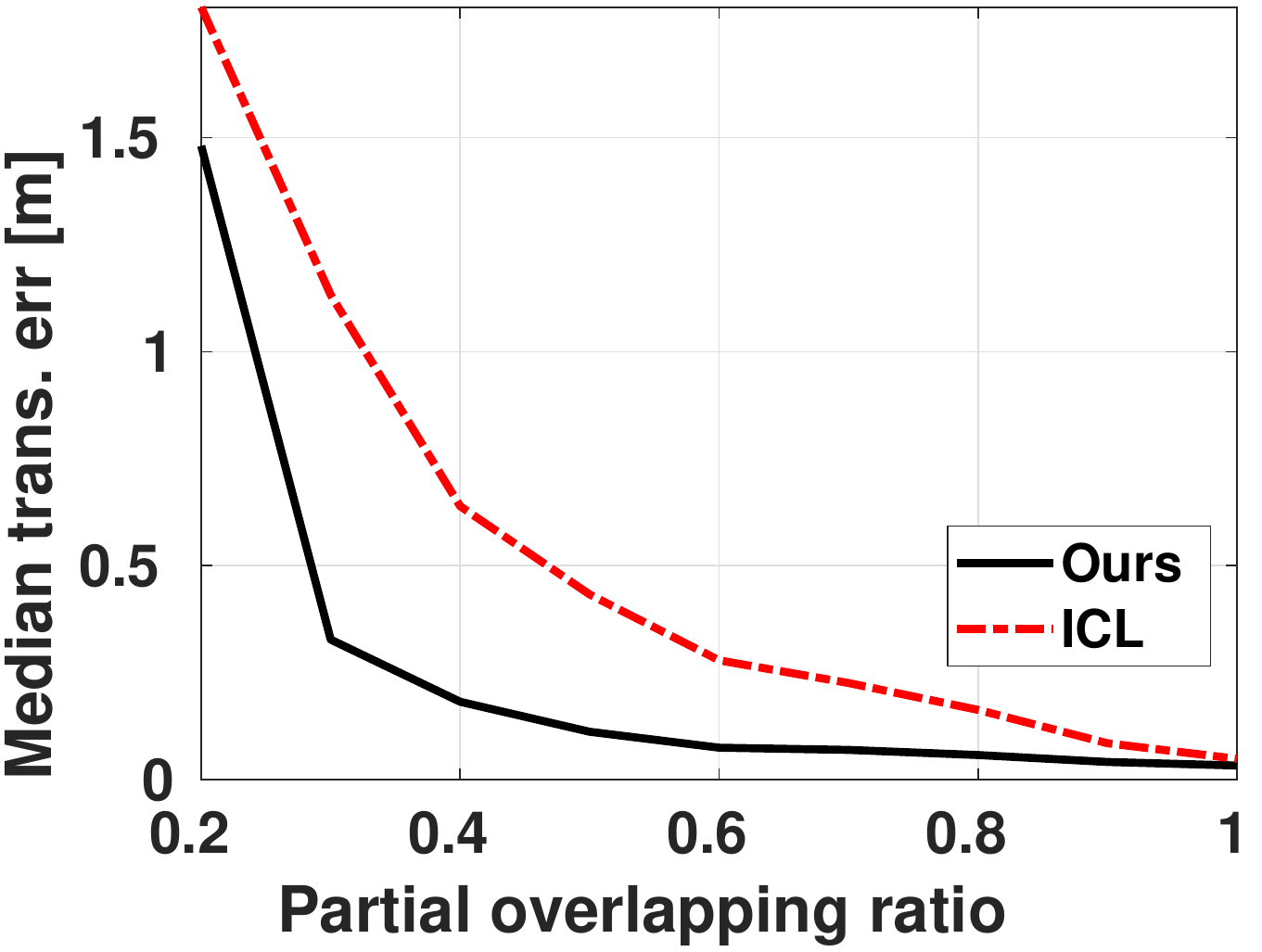}
\end{center}
\caption{\it Median rotation  (Left) and translation error (Right)  on  the  Structured3D (Top-row) and Semantic3D (Bottom-row),  with respect to increasing level of partial overlapping ratio.}
% \vspace{10pt}
\label{fig:median_err_partial_structure3Dsemantic3D}
\end{figure}

\paragraph{Robustness to overlapping ratios.} We test the  robustness of our network to overlapping ratios for partial-to-partial registration. Overlapping ratios are set within $[0.2, 1]$, where overlapping ratio at $1$ means source and target line reconstructions have one-to-one line match. For overlapping ratios smaller than $0.2$, both our method and ICL fail to estimate meaningful poses. Note we do not re-train networks. The median rotation and translation errors with respect to increasing levels of overlapping ratio are given in Figure \ref{fig:median_err_partial_structure3Dsemantic3D}. Our method outperforms ICL on the Structured3D and Semantic3D datasets.

More experiments on the effectiveness of regressing line-wise matching prior and the robustness of our network to noises are given in the appendix.

% \begin{figure}
% \begin{center}

% \end{center}
% \caption{Median rotation  (Left) and translation error (right)  on  the  Semantic3D dataset,  with respect to increasing level of partial overlapping ratio.}
% % \vspace{10pt}
% \label{fig:median_err_partial_semantic3D}
% \end{figure}

\begin{figure}
\begin{center}
\includegraphics[width=0.23\textwidth]{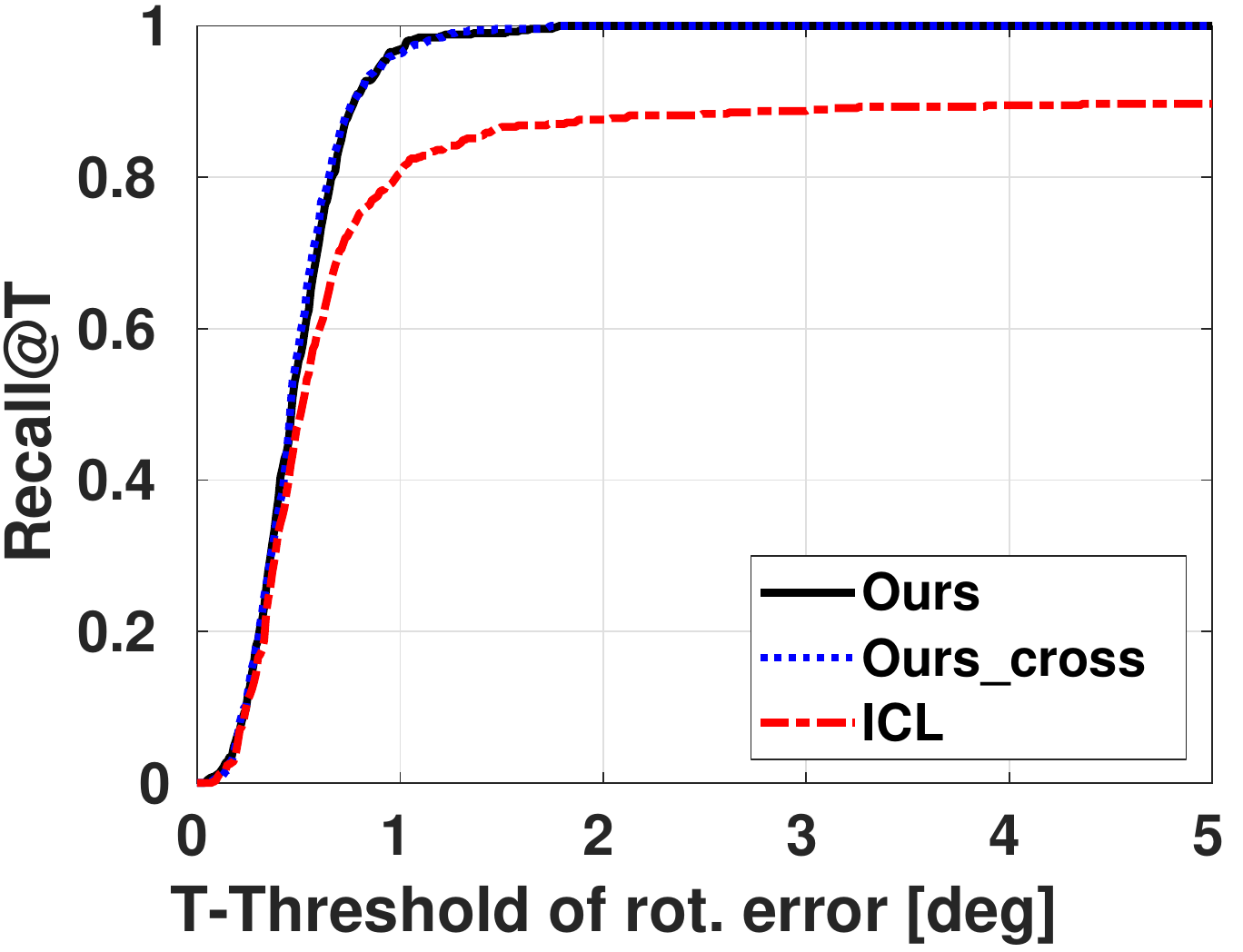}
\includegraphics[width=0.23\textwidth]{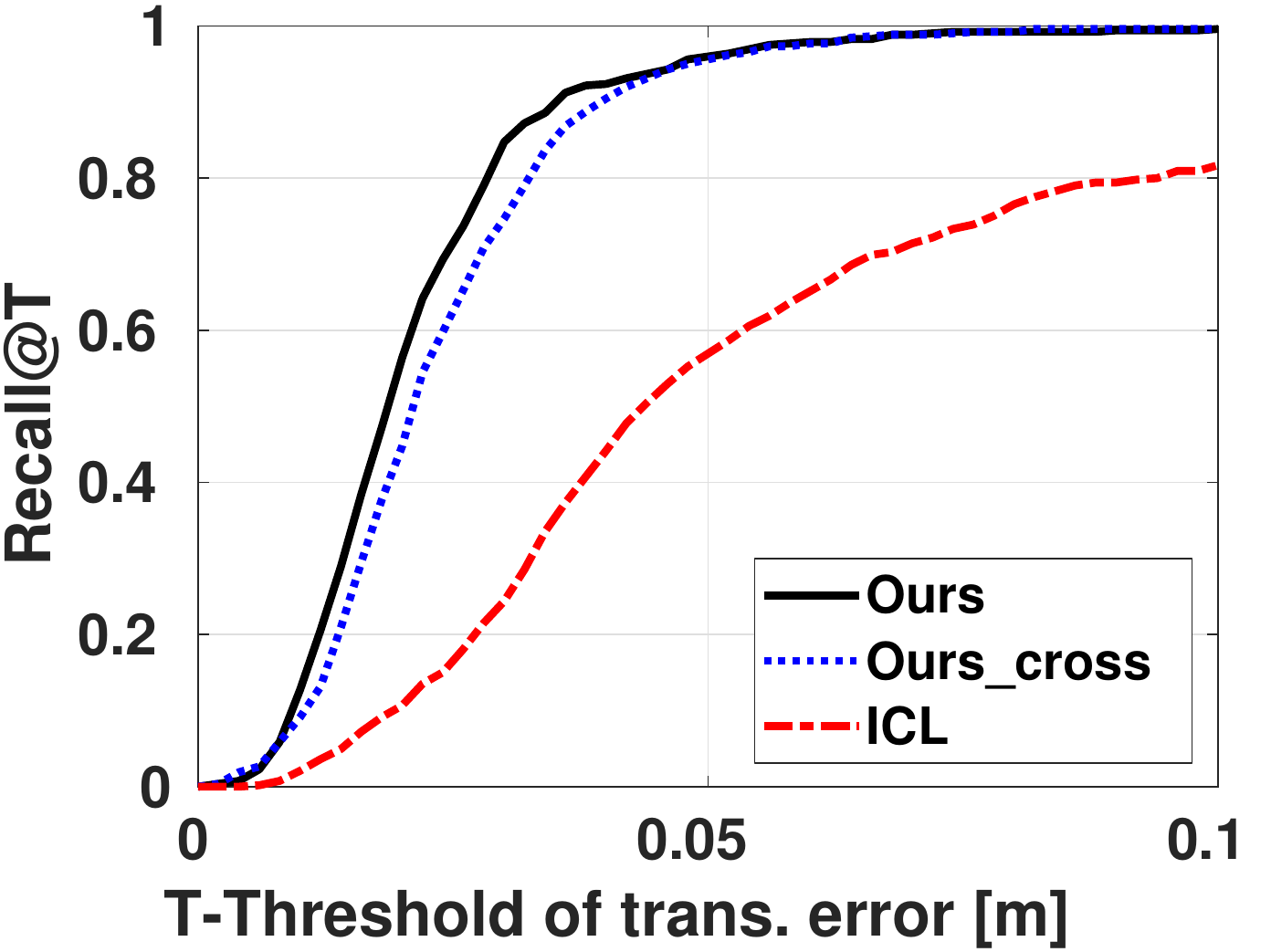}
\includegraphics[width=0.23\textwidth]{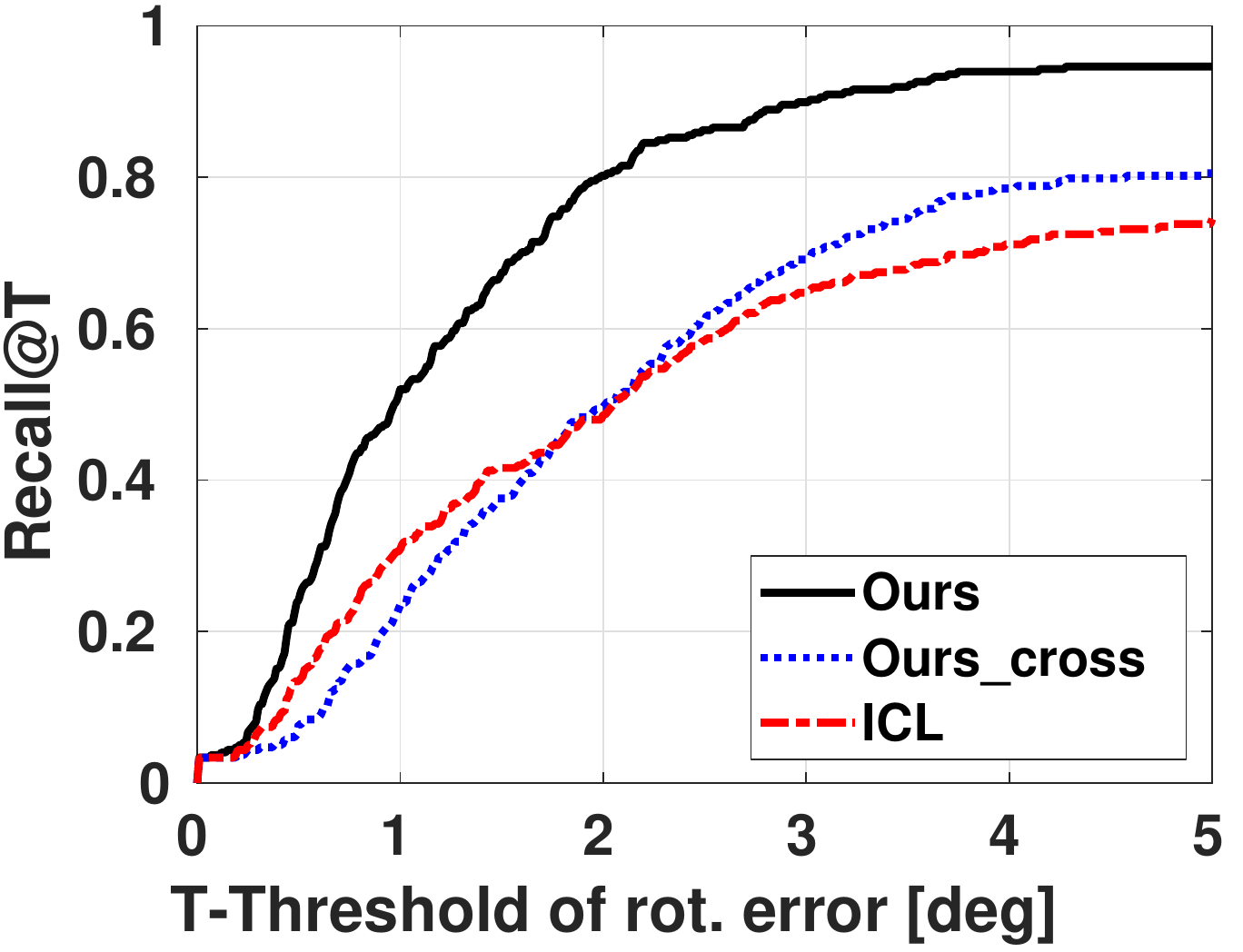}
\includegraphics[width=0.23\textwidth]{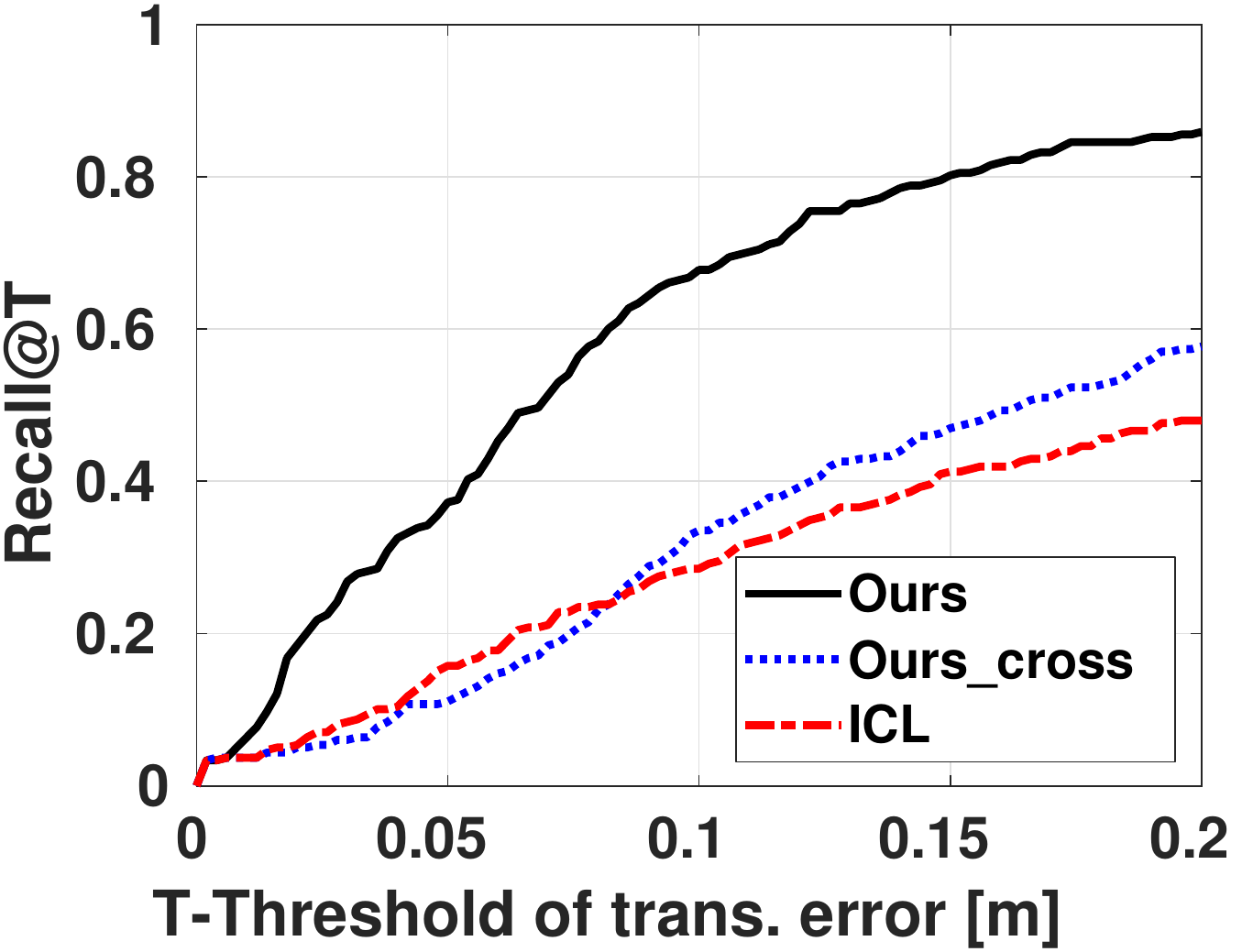}
\end{center}
\caption{\it Recall of rotation (Left) and translation (Right) on the Structured3D (Top-row) and Semantic3D (Bottom-row), with respect to an error threshold. Ours\_cross denotes using network trained on one  dataset to test on the other.}
% \vspace{10pt}
\label{fig:median_err_generalization_structure3Dsemantic3D}
\vspace*{-10pt}
\end{figure}

\paragraph{Generalization ability.} We test the generalization ability of our network via cross-dataset validation, \ie, using a network trained on the Structured3D to test its performance on the Semantic3D, and vice versa. The comparison of recall performance is given in Figure~\ref{fig:median_err_generalization_structure3Dsemantic3D}.~For the Structured3D dataset, our network trained on the Semantic3D dataset get a similar recall performance as the network trained on the Structured3D dataset, both outperforming the ICL method. For the Semantic3D dataset, though there is a recall performance drop for our  cross-dataset trained network, its performance is comparable to the ICL method.

% \begin{figure}
% \begin{center}
% \includegraphics[width=0.23\textwidth]{Figure/semantic3Drot_cmp_generalization.pdf}
% \includegraphics[width=0.23\textwidth]{Figure/semantic3Dtrans_generalization.pdf}
% \end{center}
% \caption{Recall of rotation (Left) and translation (right) on the semantic3D dataset, with respect to an error threshold. Ours\_cross denotes using network trained on the  Structure3D dataset to test on the Semantic3D dataset.}
% % \vspace{10pt}
% \label{fig:median_err_generalization_semantic3D}
% \end{figure}

\begin{figure}
\begin{center}
\includegraphics[width=0.23\textwidth]{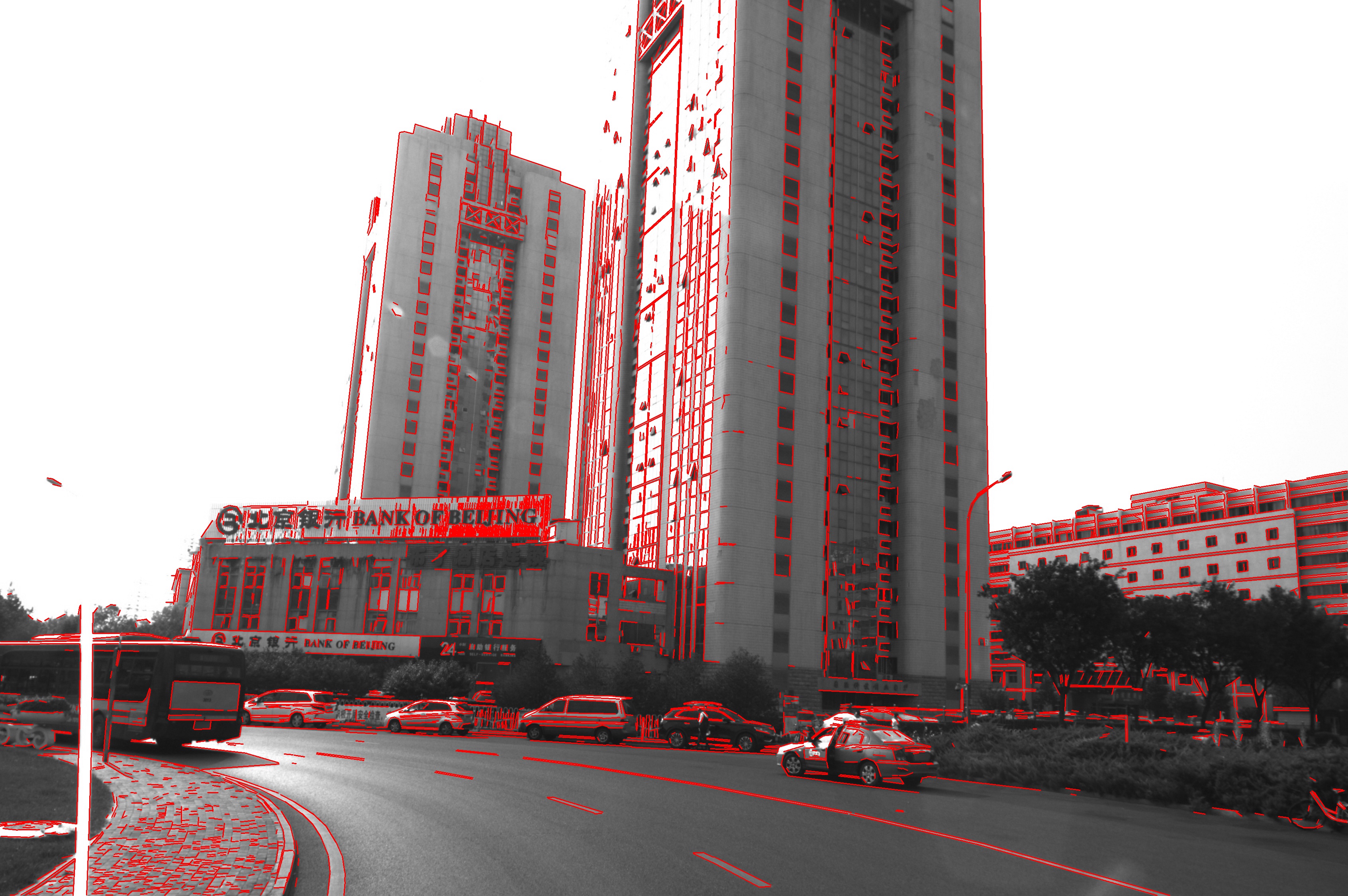}
\includegraphics[width=0.23\textwidth]{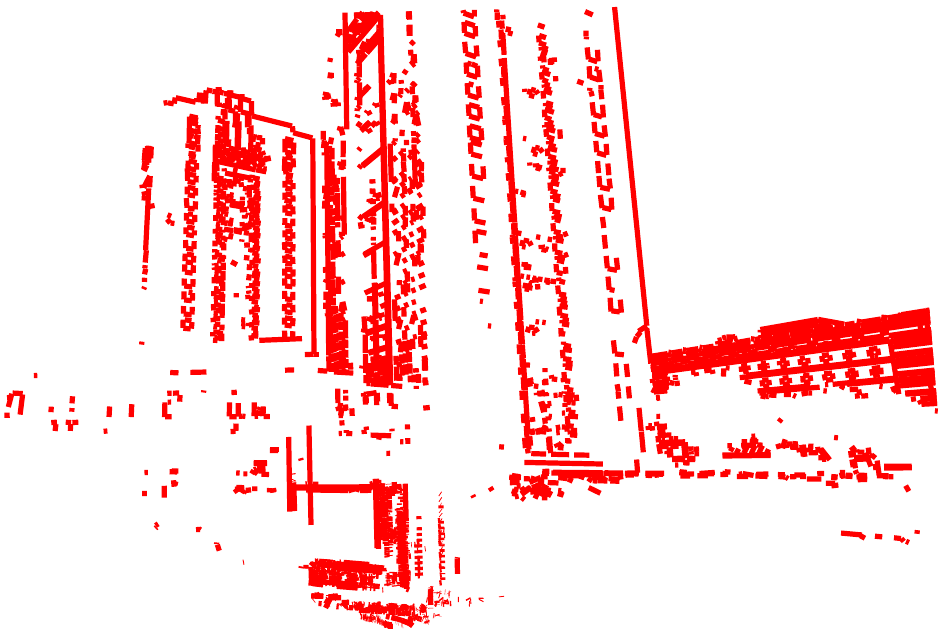}
\end{center}
\caption{\it Sample 2D (Left) and 3D (Right) line segments from the Apollo dataset.}
\vspace*{-10pt}
\label{fig:apollo_sample_segments}
\end{figure}

\paragraph{Real-world line-based visual odometry.}
In this experiment, we demonstrate the effectiveness of our method for a real-world application of registering 3D line reconstructions, \ie visual odometry. We randomly select one sequence (road02\_ins) from the apollo repository \cite{apollo_bib}, and it contains $5123$ frames. For each frame, we first use a fast line segment detector \cite{akinlar2011edlines} to obtain 2D line segments, and then use the provided depth map (from Lidar) to fit 3D lines \cite{he2018incremental}. The median number of 3D lines is $1975$. Sample 2D and 3D line segments are given in Figure \ref{fig:apollo_sample_segments}.

\begin{figure}
\begin{center}
\includegraphics[width=0.22\textwidth]{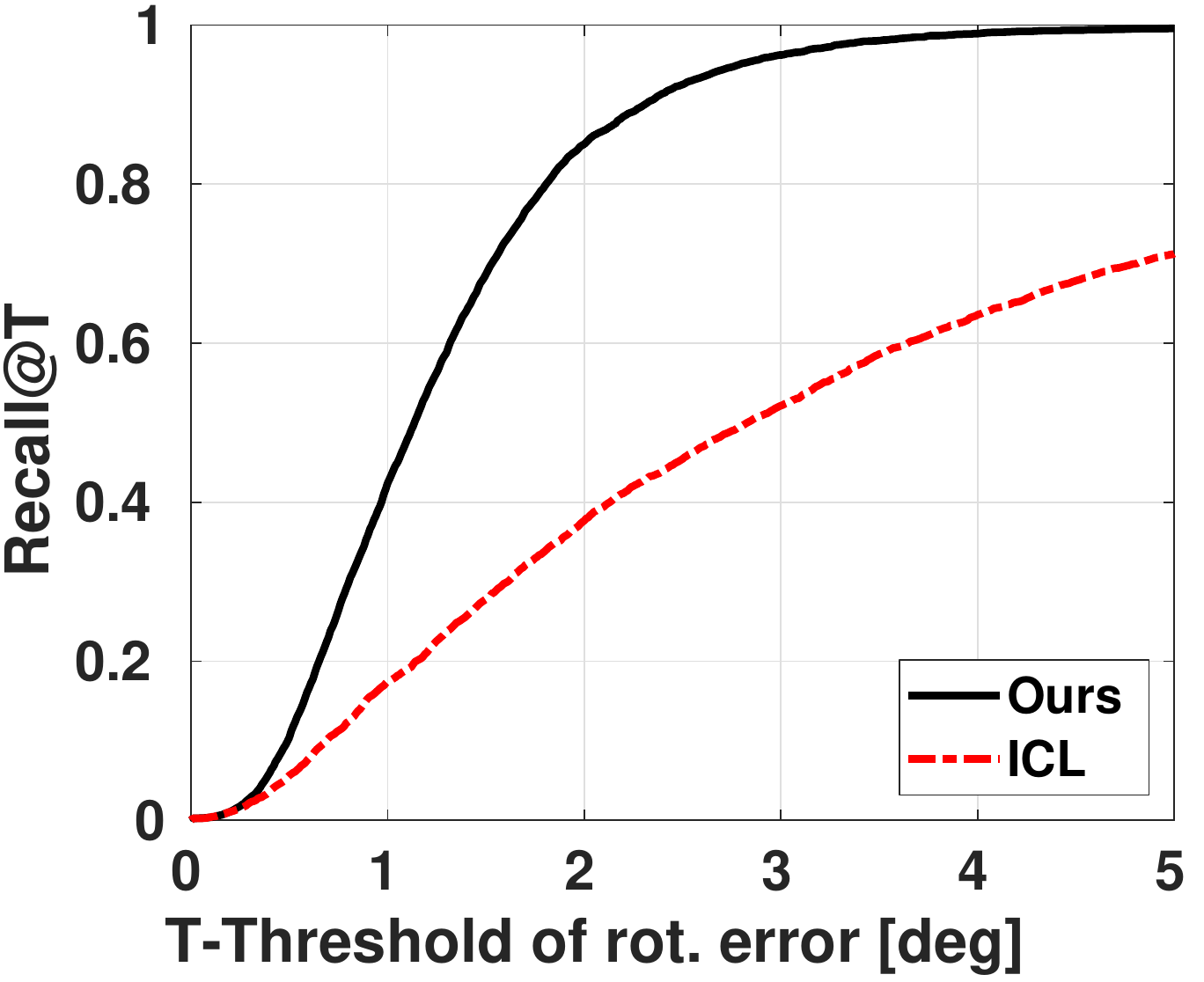}
\includegraphics[width=0.24\textwidth]{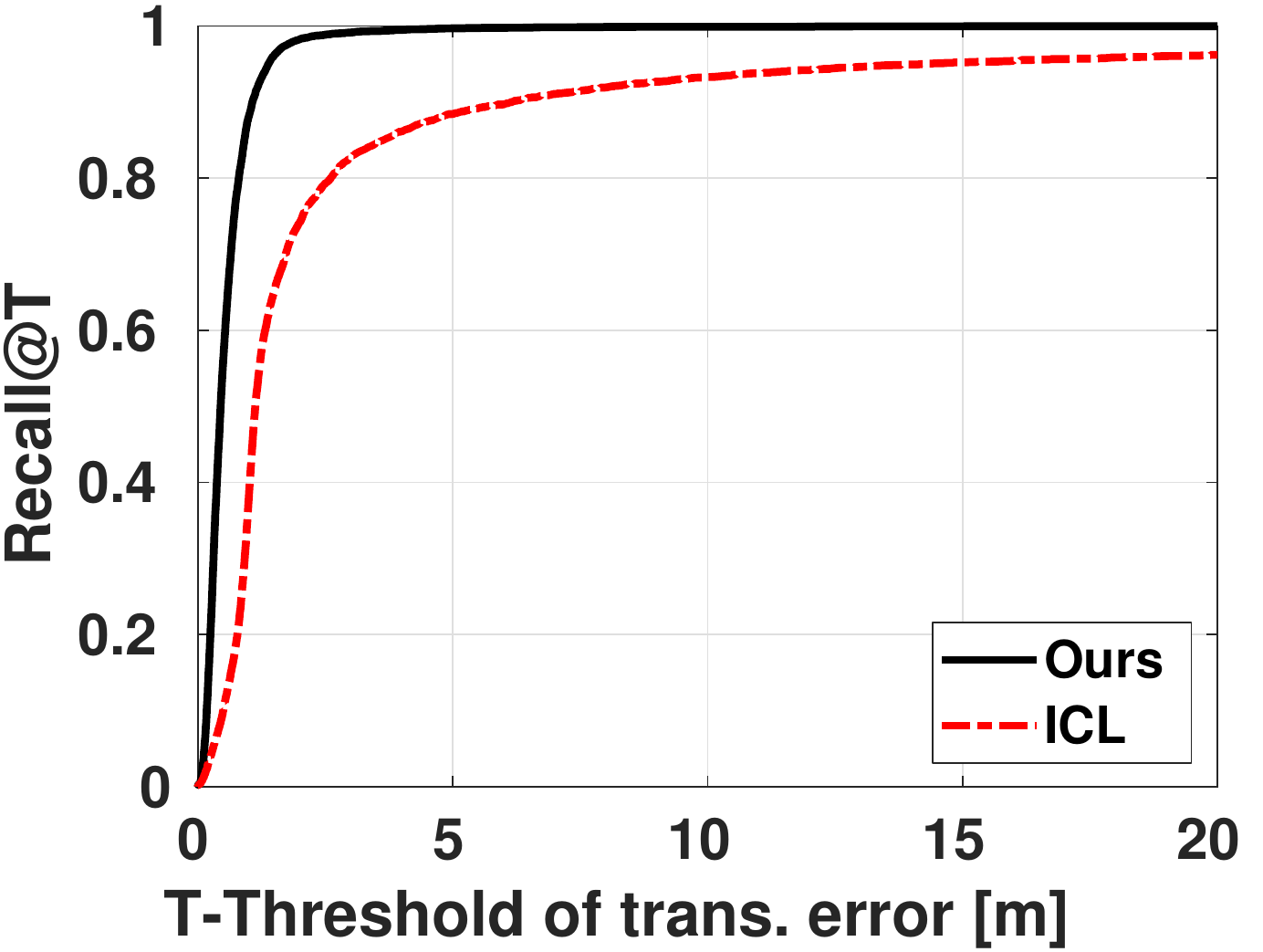}
\end{center}
\caption{\it Recall of rotation (Left) and translation (Right) on the Apollo dataset, with respect to an error threshold. We directly use model trained on Semantic3D dataset without fine-tuning.}
% \vspace{10pt}
\label{fig:apollo_vo}
\end{figure}

Given 3D line segments for each frame, we use \plucker \ representations of line segments to compute relative poses for consecutive frames, using our  model trained on the Semantic3D dataset without fine-tuning. We compare our method with ICL, and the results are given in Figure \ref{fig:apollo_vo}.
Our method outperforms ICL with much higher rotation and translation recalls, with median rotation and translation error for our method and ICL at $1.13^{\circ}/0.44\text{m}$ and $2.83^{\circ}/1.12\text{m}$, respectively.

\paragraph{Computational efficiency.}

We compare the computation time of our approach to ICL. The averaged time breakdowns for our method are $0.037\text{s}$/$0.008\text{s}$/$0.327\text{s}$ (Structured3D), $0.036\text{s}$/$0.024\text{s}$/$0.335\text{s}$ (Semantic3D) for feature extraction, matching and RANSAC pose estimation, respectively.  The averaged time for ICL is $0.055\text{s}$/$0.144\text{s}$ for Structured3D and Semantic3D, respectively. The most time-consuming part of our method is the RANSAC pose estimation step, which is programmed in Python. Interested readers can speed-up the time-consuming RANSAC iterations in C++.

\section{Conclusion}
In this paper, we have proposed the first end-to-end trainable network for solving the problem of aligning two partially-overlapped 3D line reconstructions in Euclidean space. The key innovation is to directly learn line-wise features by parameterizing lines as $6$-dim \plucker \ coordinates and respecting line geometry during feature extraction. We use these line-wise features to establish line matches via a global matching module under the optimal transport framework. The high-quality line correspondences found by our method enable to apply a $2$-line minimal-case RANSAC solver to estimate the 6-DOF pose between two line reconstructions. Experiments on three benchmarks show that the new method significantly outperforms baselines (iterative closet lines and direct regression) in terms of registration accuracy.

% \newpage
\section{Appendix}
In this appendix, we give omitted contents (for space reasons) indicated in the main paper. We present details of baseline methods, and provide more experimental results.

\begin{figure*}
\begin{center}
  \includegraphics[width=0.3\textwidth]{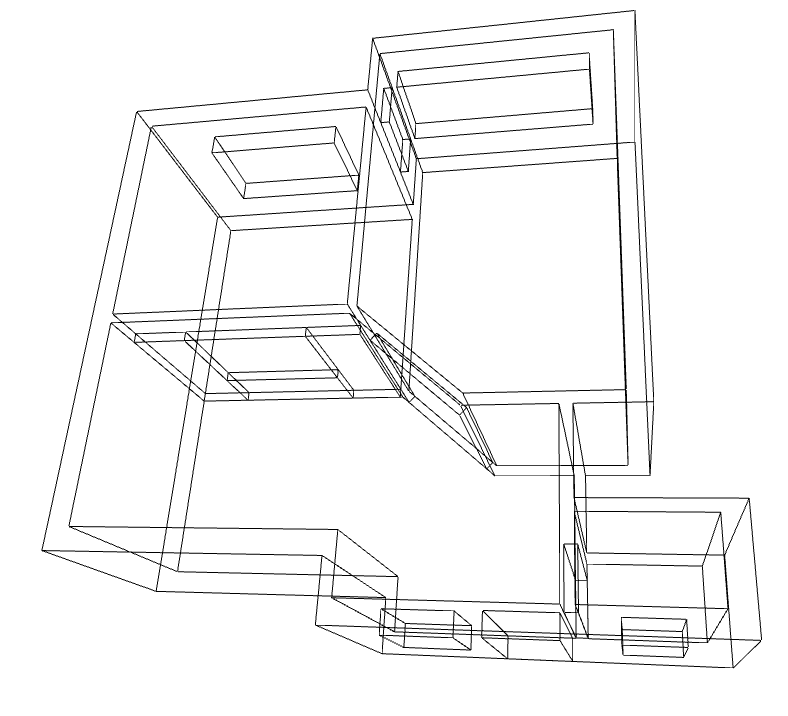}\hfill
  \includegraphics[width=0.3\textwidth]{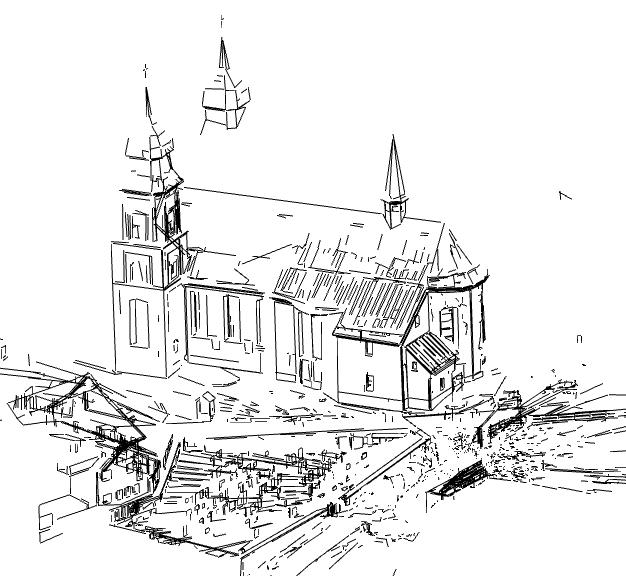}\hfill
  \includegraphics[width=0.3\textwidth,angle=180,origin=c]{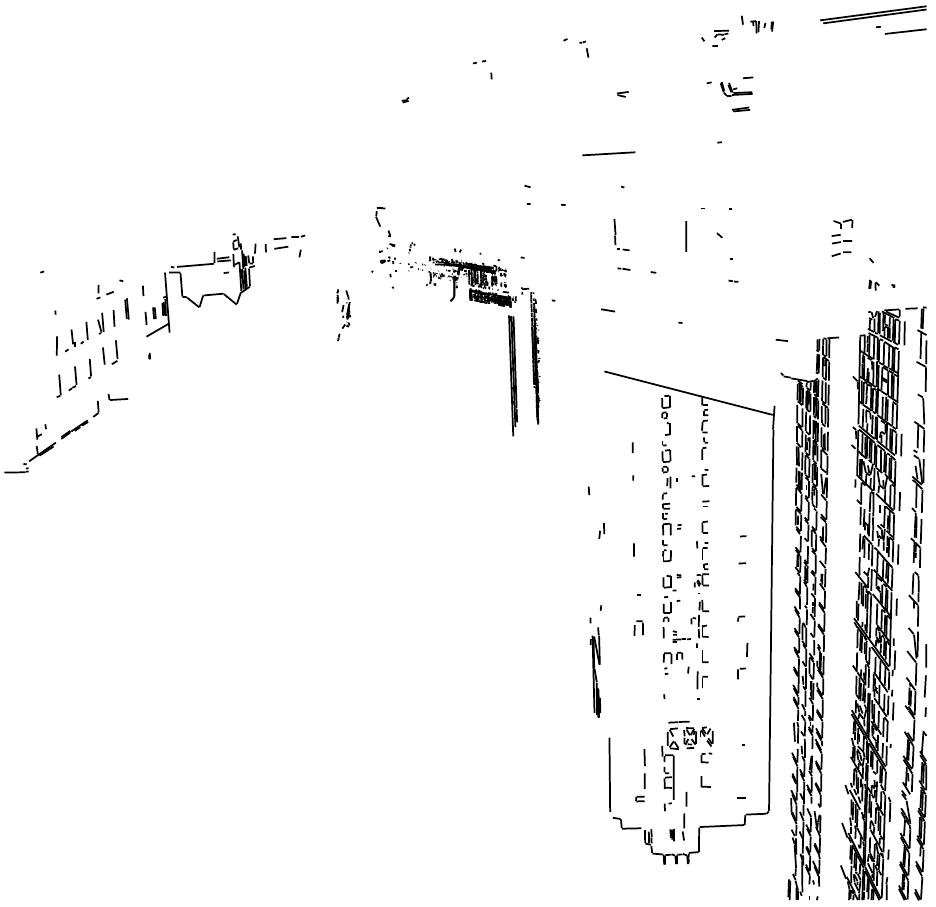}
\end{center}
\caption{\it Sample 3D line reconstructions from Structured3D (Left), Semantic3D (Middle), and Apollo (Right) datasets.}
\label{fig:dataset_demo}
\end{figure*}

\paragraph{Sample 3D line reconstructions}
\figref{fig:dataset_demo} shows sample 3D lines from the Structured3D \cite{Structured3D}, Semantic3D \cite{hackel2017isprs} and Apollo \cite{apollo_bib} datasets.

\paragraph{Iterative Closest Line (ICL).}
ICL exactly follows the pipeline of ICP (Iterative Closest Point), and the algorithm steps are:

\begin{enumerate}
    \item For each line from the source line set $L_\mathcal{S}$, find the closest line in the target line set $L_\mathcal{T}$. $L_2$ distances on $6$-dim \plucker \ coordinates of lines are employed here.
    \item Given line-to-line correspondences, estimate the relative rotation and translation using the method proposed in the pose estimation section (Section 3.4 in the main paper)
    \item Transform the source line set $L_\mathcal{S}$ using the estimated rigid transformation.
    \item Iterate the above process until stopping conditions are satisfied. We use standard stopping conditions: 1) the maximum number of iterations is reached ($100$); 2) relative change of the $L_2$ distance between $L_\mathcal{S}$ and $L_\mathcal{T}$  is sufficiently small. 
\end{enumerate}

\paragraph{Regression.} This baseline does not estimate line-to-line matches. It directly regresses a relative pose to align $L_\mathcal{S}$ to $L_\mathcal{T}$.

We first trim our network, and obtain line-wise features from the output of discriminative feature embedding  layer $^{(T)}\mathbf{f}_{\bel_i}$ and $^{(T)}\mathbf{f}_{\bel_j^{'}}$ for $L_\mathcal{S}$ and $L_\mathcal{T}$, respectively.

Line-wise features from $L_\mathcal{S}$ and $L_\mathcal{T}$ are globally max-pooled to obtain a global feature vector $\mathbf{f}_{L_\mathcal{S}}$ and $\mathbf{f}_{L_\mathcal{T}}$, respectively. $\mathbf{f}_{L_\mathcal{S}}$ and $\mathbf{f}_{L_\mathcal{T}}$ are concatenated, and passed to a MLP block MLP($256,128,128,64,64,7$) to regress a $4$-dim quaternion and $3$-dim translation. We post $L_2$ normalize the quaternion, and ensure the first component of the quaternion to be greater than $0$.

Given regressed relative pose, we use the pose regression loss to train the network:
\begin{equation}
    \mathcal{L}_\text{reg} = \left \| \mathbf{t}_\text{gt}-\mathbf{t} \right \|_2+\left \| \mathbf{q}_\text{gt}-\mathbf{q} \right \|_2,
\end{equation}
where $\mathbf{t}_\text{gt}$ and $\mathbf{t}$ are the ground-truth  and estimated translation, respectively. $\mathbf{q}_\text{gt}$ and $\mathbf{q}$ are the ground-truth  and estimated quaternion, respectively.

\begin{figure}
\begin{center}
\includegraphics[width=0.23\textwidth]{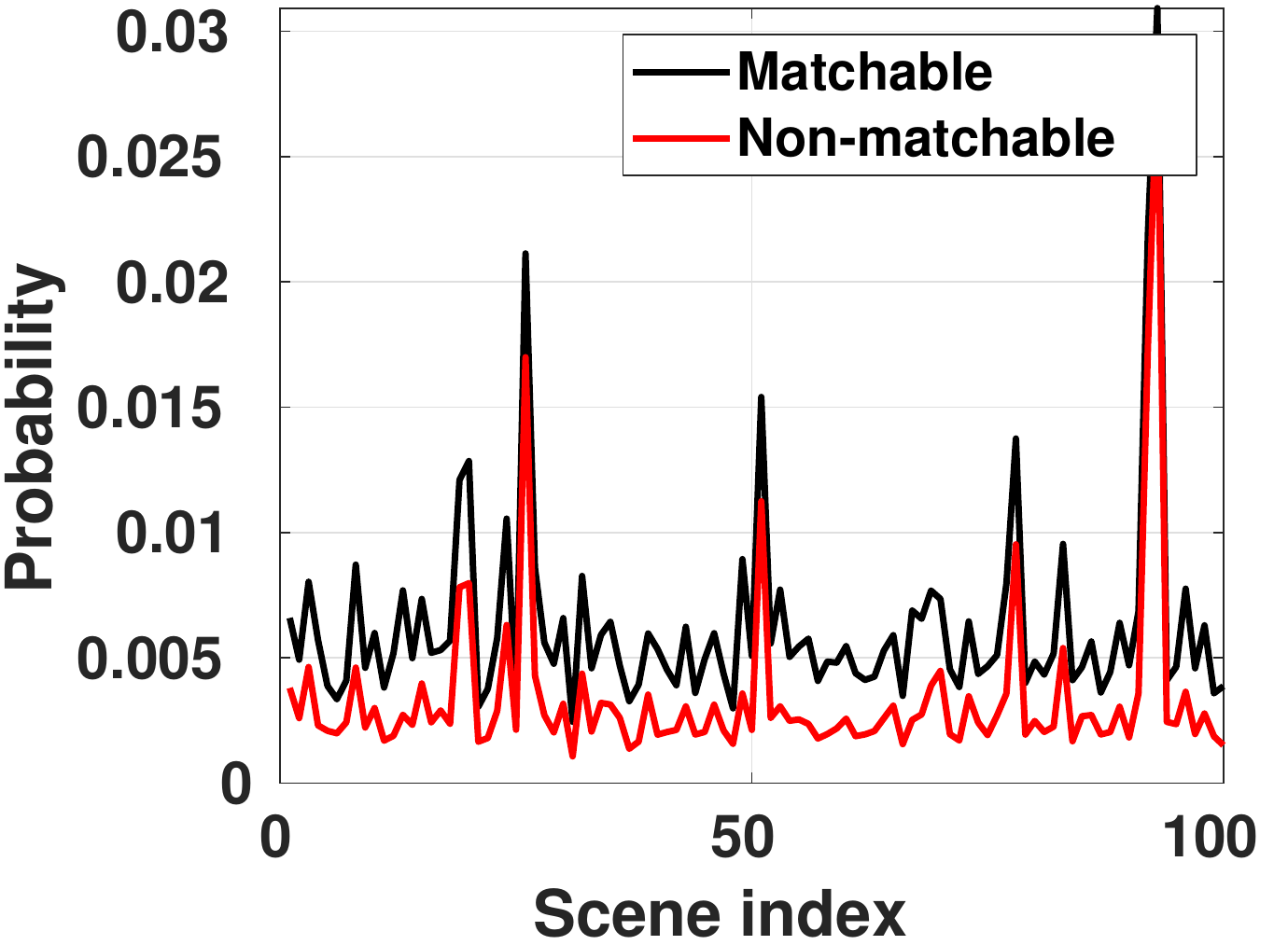}
\includegraphics[width=0.23\textwidth]{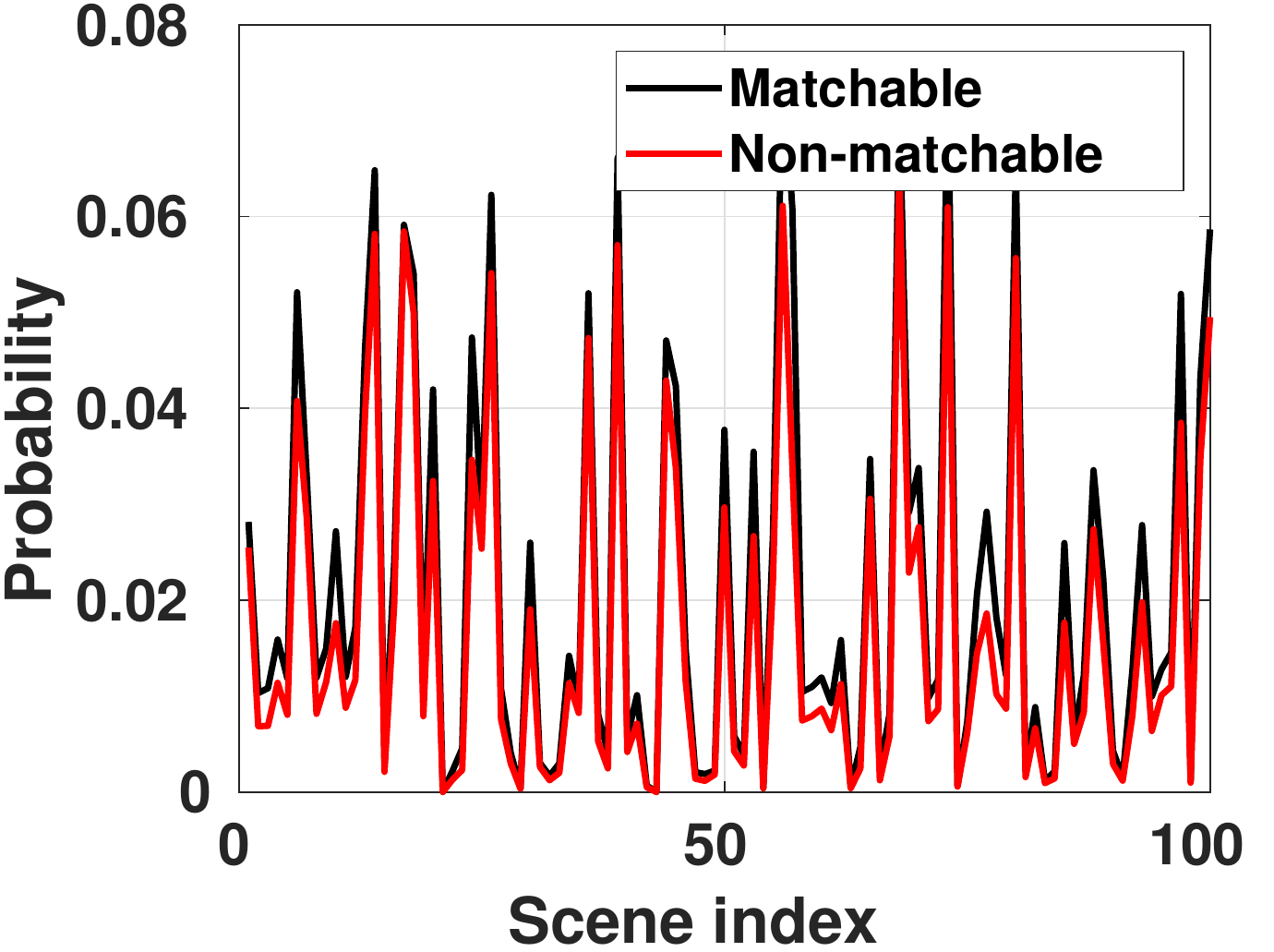}
\end{center}
\caption{\it The averaged probabilities on the  Structure3D (Left) and Semantic3D (Right) dataset for matchable and non-matchable lines. For better visualization, the first 100 scenes for each dataset are given. }
% \vspace{10pt}
\label{fig:prob_regress}
\end{figure}

\paragraph{The effectiveness of regressing line-wise matching prior}
We validate the effectiveness of regressing line-wise matching prior (Eq.~(5) in the main paper) by looking at regressed probabilities. For source and target line reconstructions of each scene, we calculate the averaged matching probabilities of matchable and non-matchable lines using ground-truth labels, and the result is given in Figure \ref{fig:prob_regress}. For almost all scenes, the estimated probability of matchable lines outweighs the probability of non-matchable lines, justifying the effectiveness of our matchability regression module. 

\begin{figure}
\begin{center}
\includegraphics[width=0.23\textwidth]{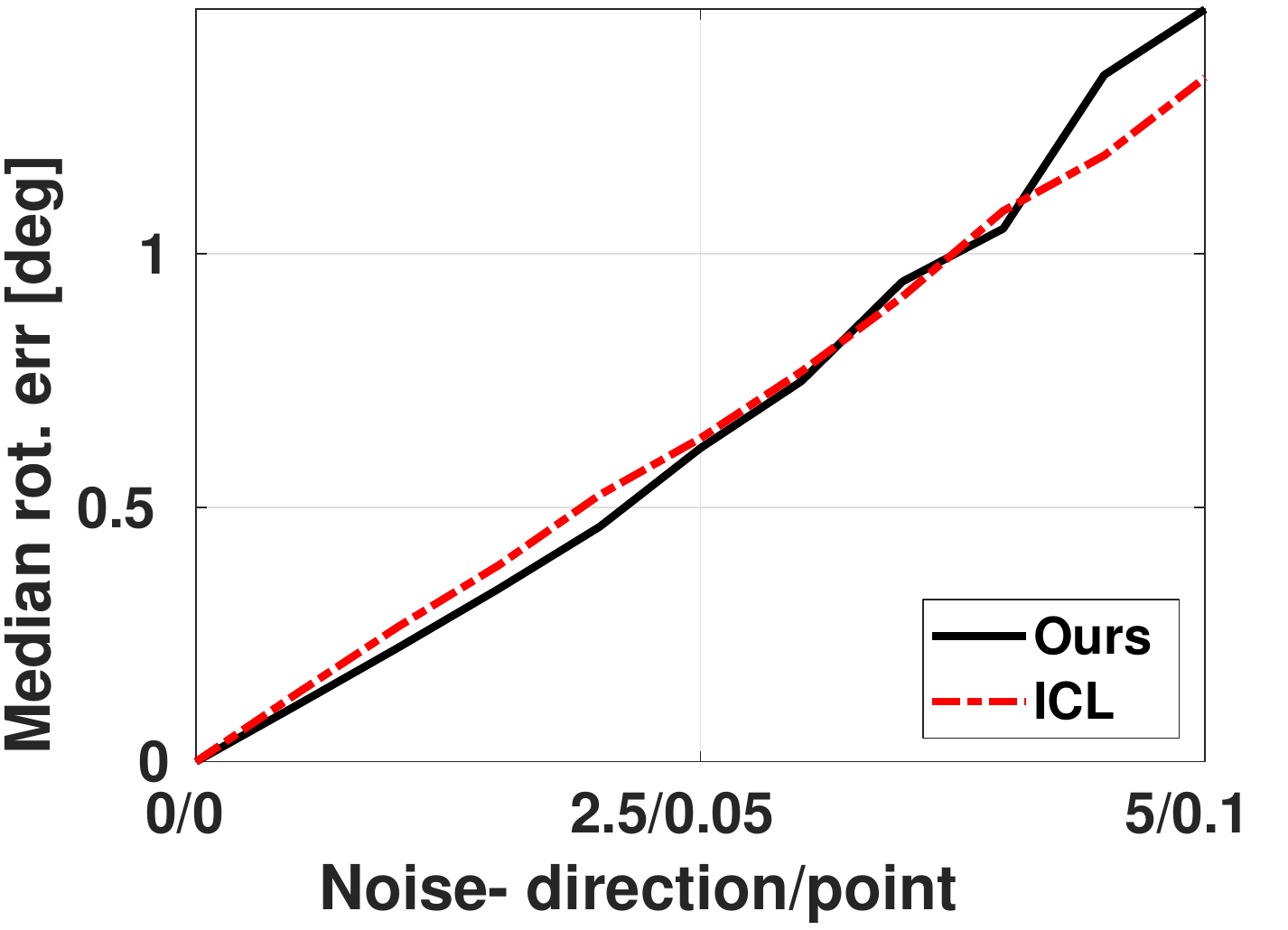}
\includegraphics[width=0.23\textwidth]{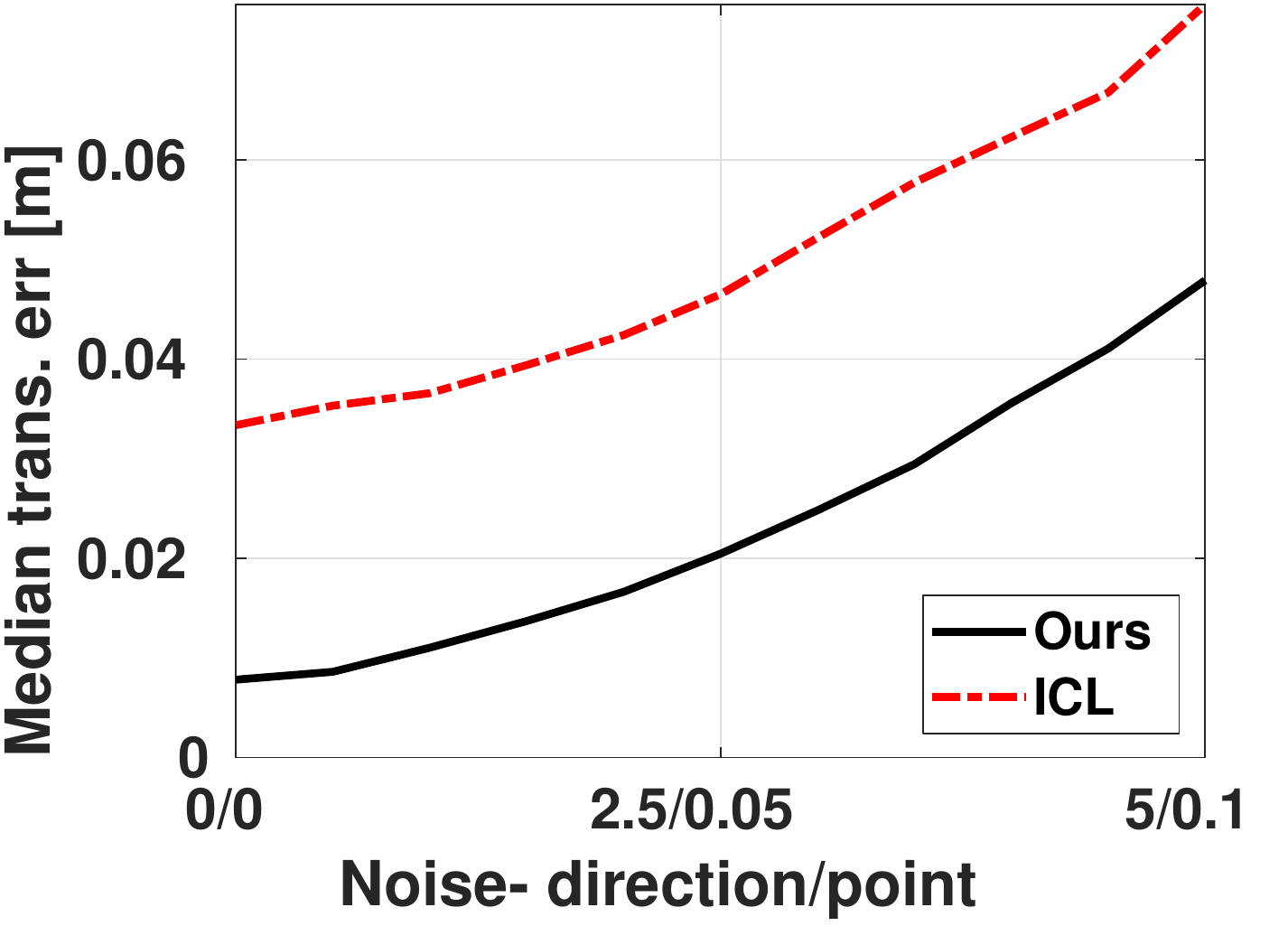}
\includegraphics[width=0.23\textwidth]{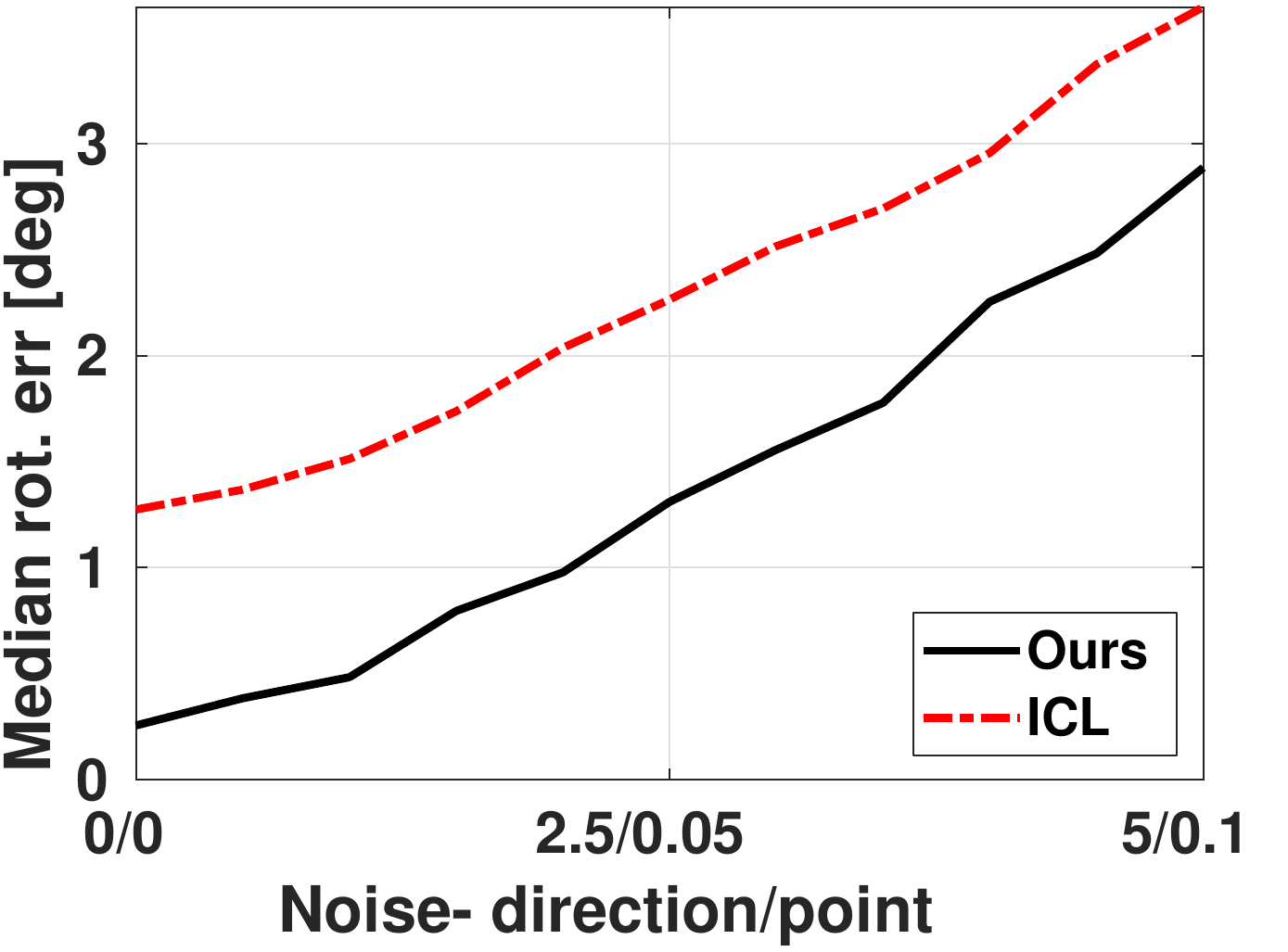}
\includegraphics[width=0.235\textwidth]{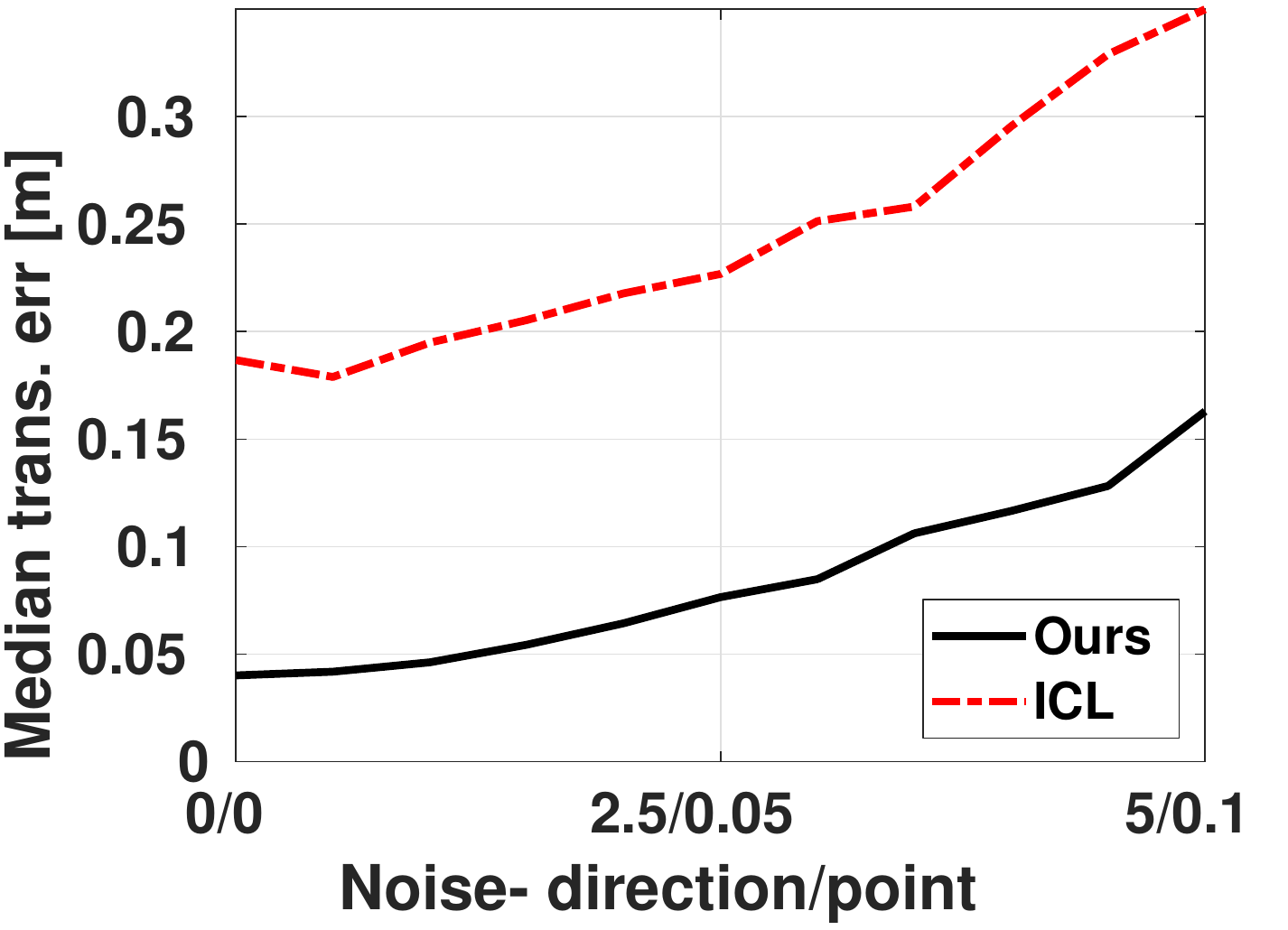}
\end{center}
\caption{\it Median rotation  (Left) and translation error (Right)  on  the  Structured3D (Top-row) and Semantic3D (Bottom-row) dataset,  with respect to increasing level of noise.}
% \vspace{10pt}
\label{fig:median_err_noise_structure3Dsemantic3D}
\end{figure}

\paragraph{Robustness to noises}
Though we have already added noise to line reconstructions (see paragraph partial-to-partial registration in the main paper), to further evaluate the robustness of our networks to noise, we perform experiments with different levels of Gaussian noise for both source and target line reconstructions. A line direction and footprint is perturbed by a random rotation and translation, respectively.  Rotation angles and footprint translations are sampled from $\mathcal{N}(0,\sigma_a)$ and $\mathcal{N}(0,\sigma_f)$, respectively. We construct consecutive tuples for $(\sigma_a, \sigma_f)$, ranging from $(0,0)$ to $(5,0.1)$.   Note that we do not re-train networks. The median rotation and translation error with respect to an increasing level of noise is given in Figure \ref{fig:median_err_noise_structure3Dsemantic3D}. Our method gets similar median rotation errors as ICL on the Structured3D dataset, while outperforms it for translation error. Our method outperforms ICL on the Semantic3D dataset.

{\small
\bibliographystyle{ieee_fullname}
\bibliography{egbib}
}
\end{document}